\documentclass[lettersize,journal]{IEEEtran}
\usepackage{amsmath,amsfonts}
\usepackage{algorithm}
\usepackage{algpseudocode}
\usepackage{array}
\usepackage{textcomp}
\usepackage{stfloats}
\usepackage{url}
\usepackage{verbatim}

\makeatletter
\def\whline#1{%
	\noalign{\ifnum0=`}\fi\hrule \@height #1 \futurelet
	\reserved@a\@xhline}

\usepackage{paralist}

\usepackage{graphicx}
\usepackage{cite}
\usepackage{threeparttable}
\usepackage{makecell}
\usepackage{multirow}
\usepackage{footnote}
\usepackage{tablefootnote}
\usepackage{float}
\usepackage[numbers]{natbib}
\usepackage{colortbl}
\usepackage{array}
\usepackage{caption}
\usepackage{booktabs}
\usepackage{ragged2e}
\usepackage{setspace}
\usepackage{enumitem}
\usepackage{bbm}
\usepackage{subcaption}
\usepackage{float}
\usepackage[bookmarksnumbered=true, pagebackref=false]{hyperref}
\definecolor{seagreen}{rgb}{0.1,0.92,0.21}
\definecolor{darkgreen}{rgb}{0.21,0.82,0.11}
\definecolor{deepblue}{rgb}{0.12,0.25,0.62}
\definecolor{darkred}{rgb}{0.92,0,0.12}

\hypersetup{colorlinks,
	citecolor=seagreen,
	filecolor=[rgb]{0.6,0.0,0.4},
	linkcolor=darkred,
	urlcolor=[rgb]{0.0,0.3,0.7},
}

\begin{document}

\title{Online Writer Retrieval with Chinese Handwritten Phrases: A Synergistic Temporal-Frequency Representation Learning Approach}

\author{Peirong Zhang,~Lianwen Jin$^*$~\IEEEmembership{Member,~IEEE}
\thanks{Manuscript received March 16, 2024; revised October 7, 2024. This research is supported by National Key Research and Development Program of China (2022YFC3301703).}
\thanks{Peirong Zhang and Lianwen Jin are with the School of Electronic and Information Engineering, South China University of Technology, Guangzhou, China.}
\thanks{$^*$ Corresponding author.}
}

\markboth{IEEE TRANSACTIONS ON INFORMATION FORENSICS AND SECURITY,~Vol.~19, 2024}%
{Shell \MakeLowercase{\textit{et al.}}: A Sample Article Using IEEEtran.cls for IEEE Journals}


\maketitle

\begin{abstract}
Currently, the prevalence of online handwriting has spurred a critical need for effective retrieval systems to accurately search relevant handwriting instances from specific writers, known as online writer retrieval. Despite the growing demand, this field suffers from a scarcity of well-established methodologies and public large-scale datasets. This paper tackles these challenges with a focus on Chinese handwritten phrases. First, we propose DOLPHIN, a novel retrieval model designed to enhance handwriting representations through synergistic temporal-frequency analysis. For frequency feature learning, we propose the HFGA block, which performs gated cross-attention between the vanilla temporal handwriting sequence and its high-frequency sub-bands to amplify salient writing details. For temporal feature learning, we propose the CAIR block, tailored to promote channel interaction and reduce channel redundancy. Second, to address data deficit, we introduce OLIWER, a large-scale online writer retrieval dataset encompassing over 670,000 Chinese handwritten phrases from 1,731 individuals. Through extensive evaluations, we demonstrate the superior performance of DOLPHIN over existing methods. In addition, we explore cross-domain writer retrieval and reveal the pivotal role of increasing feature alignment in bridging the distributional gap between different handwriting data. Our findings emphasize the significance of point sampling frequency and pressure features in improving handwriting representation quality and retrieval performance. Code and dataset are available at \url{https://github.com/SCUT-DLVCLab/DOLPHIN}.
\end{abstract}


\section{Introduction}
\begin{figure}[t]
	\centering
	\includegraphics[width=\linewidth]{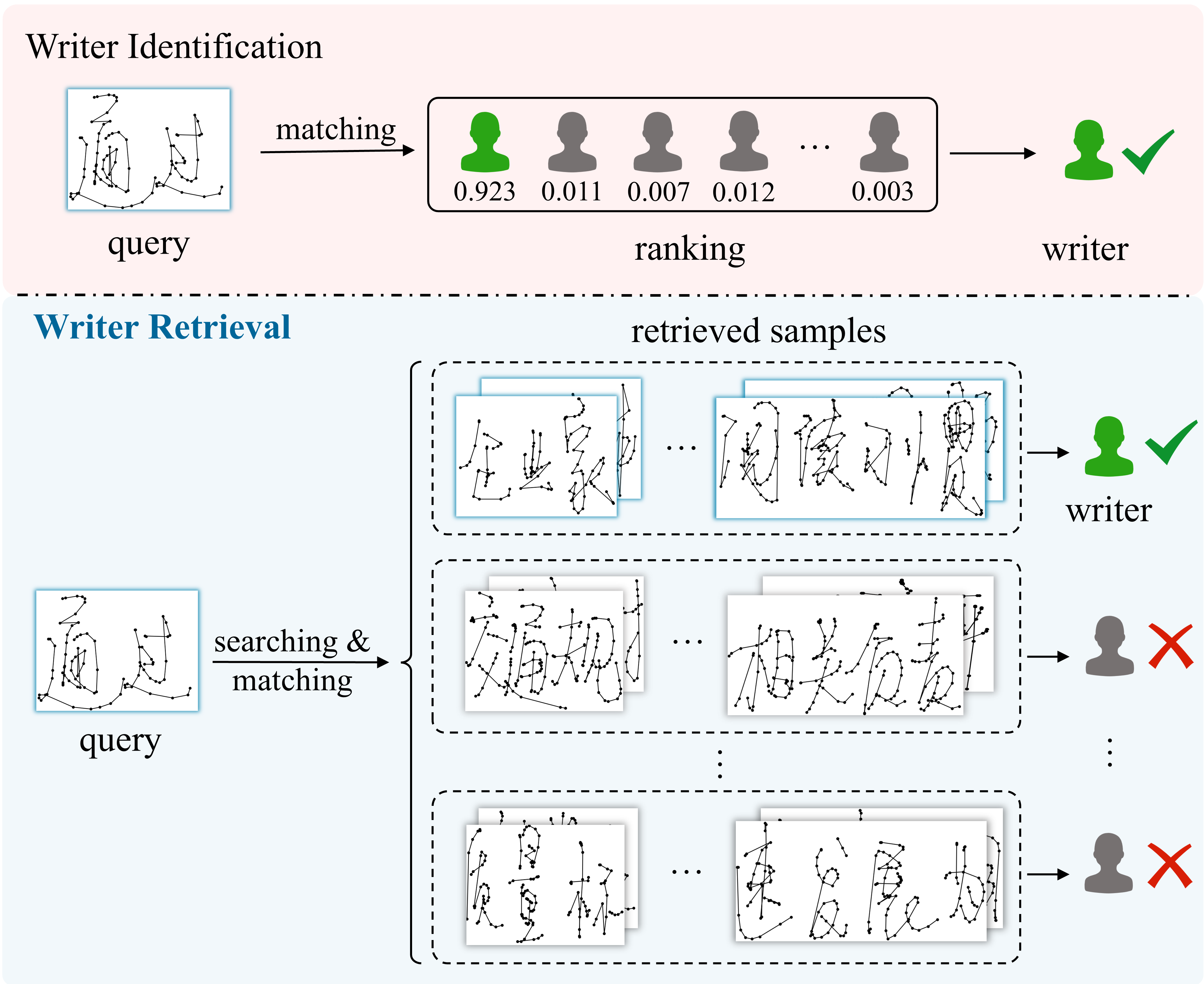}{}
	\caption{Comparison between writer identification and writer retrieval. Writer identification merely determines the authorship of handwriting without searching all relevant samples, while writer retrieval encompasses both functionalities.}
	\label{Fig::tasks}
\end{figure}

In the ever-evolving landscape of handwriting analysis and digital forensics, writer retrieval \cite{scriptorient2011icdar} has established itself as a reliable and enduring technique for decades. This process entails searching for all samples of a specific writer within a handwritten document collection. Contingent on the data nature, writer retrieval could be dichotomized into online and offline modalities. Online writer retrieval \cite{rehman2019writer} involves retrieving dynamic handwriting samples captured by devices like touchscreens and digital styluses, whereas offline writer retrieval \cite{local2012IAPR,tricnn2018keglevic} refers to searching for the text written on static images. In the digital era, the proliferation of online communication and the advancement of pen-based interfaces have led to widespread use of online handwritten data. Therefore, the demand for online writer retrieval has become increasingly paramount, such as tracking potential suspects in forensic investigations by identifying similar handwriting. This paper focuses on online writer retrieval. Notably, there is a related field known as online writer identification (Writer-ID) \cite{deepwriterid2016weixin,chen2021level}. However, Writer-ID solely focuses on determining the authorship of handwriting, while writer retrieval more broadly encompasses both the retrieval of related samples and the determination of authorship. A visual comparison of the two tasks is depicted in Fig.~\ref{Fig::tasks}.

In the context of modern online writer retrieval, retrieval systems confront several key challenges. First, driven by the fast-paced nature of digital communication where concise and brief messages are preferable, users tend to handwrite succinct snippets rather than long-form documents to perform real-time interactions, such as signing signatures or taking quick notes. Therefore, available online handwriting data in practical applications is typically short, being usually words or phrases. This presents larger difficulties for retrieval systems since phrase/word-level data contains fewer handwritten features. Second, in forensic investigation, the retrieval database usually comprises millions of handwriting samples. This is to ensure comprehensive coverage, particularly in high-profile cases involving numerous potential suspects or widespread cybercrimes. Hence, it poses a high requirement regarding the efficiency of the retrieval system to handle a large volume of data within a reasonable timeframe. Third, writer retrieval scenarios could be categorized as open-set and closed-set. Open-set retrieval refers to obtaining similar samples given the query of unknown writers in an extra data pool, whereas closed-set retrieval aims to match the query to one of the known candidate writers and find all relevant instances. Open-set scenarios, while more flexible, pose greater challenges by tasking the systems with more demanding generalizability. A modern writer retrieval system is expected to handle both open-set and closed-set retrievals, particularly the open-set scenarios for broader and more exploratory applications. This paper primarily focuses on open-set retrieval.

While offline writer retrieval has seen advancements, online writer retrieval remains nearly uncharted, with few established methodologies and large-scale datasets. To propel this shadowed field into the spotlight, we propose \textbf{DOLPHIN}, an online writer retrieval model designed to improve stylistic handwriting representation. First, we propose a High Frequency Gated Attention (HFGA) block to excavate the salient writing features through high-frequency attention. We utilize the Discrete Wavelet Transform to extract the high-frequency components of handwriting, which then undergo gated cross-attention \cite{flamingo2022nips} with the original time series. This approach harnesses the synergy between temporal and frequency domains, not only amplifying the local discriminative writing features but also reinforcing holistic feature representations. Second, we propose a Channel Activation Inverted Residual (CAIR) block to reduce channel information redundancy. CAIR incorporates a modified Inverted Residual \cite{mobilev22018sandler} with multiple channel activation techniques \cite{shufflenet2018zhang,howard2017mobilenets}, collaboratively activating the channel information flow to reduce redundant traits. In addition, we introduce a Context-Aware FPN by integrating the feature pyramid network (FPN) \cite{fpn2017lin} and the Global Context block \cite{gcnet2019iccvw}. This approach captures multi-scale features with improved contextual modeling, resulting in more comprehensive feature embedding for personal writing styles.

Subsequently, we build a large-scale retrieval dataset that fulfills the aforementioned data requirements. We introduce the \textbf{O}n\textbf{LI}ne \textbf{W}rit\textbf{E}r \textbf{R}etrieval (OLIWER) dataset, composed of 674,017 online Chinese handwritten phrases from 1,731 writers, aggregated from three public online handwriting datasets, namely CASIA-OLHWDB \cite{casiaolhwdb2011liu}, DCOH \cite{DCOH2024dlvc}, and SCUT-COUCH2009 \cite{couch20092011jin}. To acquire phrase-level (within two to five characters) handwritten data, we segment text lines from CASIA-OLHWDB and DCOH into discrete phrases and directly gather the words/phrases within this length of SCUT-COUCH2009. To the best of our knowledge, this is the first large-scale dataset tailored for online writer retrieval.

We conduct extensive open-set evaluations with four datasets, including the proposed OLIWER dataset, the segmented CASIA-OLHWDB, the segmented DCOH, and SCUT-COUCH2009. Owing to the scarcity of existing methods in online writer retrieval, we migrate some established models from other domains such as writer identification \cite{grrnn2021he}, signature verification \cite{lai2021synsig2vec}, and person re-identification \cite{cdnet2021li,cal2021rao,osnet2022zhou}, for comparisons with our proposed DOLPHIN. Experimental results demonstrate a pronounced outperformance of our model over prior methods across multiple datasets. DOLPHIN also exhibits practical superiorities, such as real-time inference speed and low computational cost, which could potentially benefit real-world applications. In addition, we conduct experiments on the similar writer identification task, in which DOLPHIN's superior performance further substantiates its effectiveness and generalizability. Furthermore, we investigate the cross-domain writer retrieval. We reveal that increasing feature alignment plays a crucial role in bridging the distribution gap across different handwriting data domains. In this process, we further discover that a higher point sampling ratio and the inclusion of pressure information in online handwritten data can significantly enhance retrieval performance.

Our main contributions include:
\begin{itemize}
	\item We propose DOLPHIN, a novel model that enhances online handwriting representation learning through coordinated temporal-frequency analysis. Standing as the core module, HFGA spotlights salient handwritten patterns through gated cross-attention between the high-frequency components and original temporal sequences. Meanwhile, CAIR is introduced to diminish channel redundancy and bolster temporal feature modeling. These innovations collaboratively enable DOLPHIN to extract informative writing features for matching and retrieval, even amidst highly variable handwriting styles.
	\item We conduct comprehensive experiments on multiple datasets to assess the efficacy of DOLPHIN. Results showcase DOLPHIN's significant performance improvements over preceding methods, especially on the most difficult mAP metric.
	\item We reveal that increasing point sampling frequency and incorporating pressure information can not only mitigate the distributional gap between distinct handwriting domains but also significantly improve model performance.
	\item We instigate a renaissance of the largely untapped field of online writer retrieval, beckoning more research attention and community endeavors to invigorate this realm. We hope the DOLPHIN model and the OLIWER dataset could serve as the groundwork for further advancements.
\end{itemize}

\section{Related Work}
\label{sec::related}
\subsection{Writer Retrieval}
Existing writer retrieval methods include codebook-based or codebook-free methods, in which codebook-based methods are dominated by Vector of Locally Aggregated Descriptors (VLAD) \cite{vlad2012tpami}, while codebook-free ones mainly rely on deep neural networks.

\subsubsection{Codebook-Based Methods}
Codebook-based methods can be more precisely categorized as manual codebook learning \cite{local2012IAPR,fvbased2013fiel,unsupervised2017chris} and auto codebook learning \cite{rasoulzadeh2022writer,netmvlad2022peer,rethis2023icdar}. The first category demands constructing codebooks with handcrafted/neural network-extracted features. Fiel and Sablatnig \cite{local2012IAPR} calculated SIFT features on handwritten images to construct a codebook, and then created an occurrence histogram for feature comparisons. They also proposed to encode SIFT features with Fisher Vectors \cite{fishervector2013sanchez} to form the codebook in \cite{fvbased2013fiel}. Christlein et al. \cite{unsupervised2017chris} trained a CNN to learn local features in the unsupervised manner, and used the CNN activation features to create codebooks via VLAD. Afterward, Arandjelovic et al. \cite{netvlad2016cvpr} proposed the NetVLAD layer, a parametrized, trainable, and pluggable version of VLAD that enables autonomous codebook learning. Rasoulzadeh and BabaAli \cite{rasoulzadeh2022writer} placed the NetVLAD layer at the end of a ResNet-20 backbone to learn the codebook embedding. Peer et al. \cite{netmvlad2022peer} proposed NetMVLAD by incorporating multiple vocabularies into vanilla VLAD. In their following work \cite{rethis2023icdar}, they improved NetVLAD to NetRVLAD by removing original pre- and intra-normalization layers, and also placed NetRVLAD at the tail of a ResNet.

\subsubsection{Codebook-Free Methods}
Atanasiu et al. \cite{scriptorient2011icdar} extracted 10 perceptual features of handwriting from the probability density function of the contour orientations. Fiel and Sablatnig \cite{cnn2015fiel} used a pure CNN to extract features of each handwriting patch, and average vectors of all patches to generate ultimate embeddings for retrieval. Keglevic et al. \cite{tricnn2018keglevic} proposed to use a triplet CNN to encode image patches and perform metric learning in tandem. They used VLAD for post-CNN feature encoding while they did not generate any codebook. Koepf et al. \cite{writeridvit2022koepf} exploited the ViT-Lite as the feature extractor and aggregated the feature representations for retrieval.

Despite these advancements, research has predominantly focused on offline writer retrieval, leaving online writer retrieval largely unexplored. The retrieval method development is mainly restricted by the scarcity of available large-scale datasets. Although datasets like IAM-OnDB \cite{iamonline2005liwicki} and IBM\_UB\_1 \cite{ibm2013shivram} exist for online writer identification and can be used for online writer retrieval through rearrangement, they are limited in sizes, the number of users, and writing style diversities. Therefore, it puts a clear demand on large-scale online handwritten datasets to facilitate the advancement of specialized retrieval methods in this area. To this end, we propose the OLIWER dataset with both large data volume and rich writer diversity, and build our DOLPHIN model towards better writer retrieval through temporal-frequency synergistic representation learning.

\subsection{Writer Identification}
Writer identification (Writer-ID) pinpoints the determination of a single writer's identity for a particular handwriting, exhibiting a more constrained functionality compared to writer retrieval. Still, both tasks share a similar objective of analyzing handwriting to determine the writer belongings. Prior Writer-ID techniques can be grouped into two categories: statistical-based and deep neural network-based. 

\subsubsection{Statistical-Based Methods}
These methods typically include two stages: the feature extraction stage and the statistical modeling stage. For the first stage, researchers have explored hand-crafted features including allograph features \cite{tan2009allograph,tan2010individuality}, point-/stroke-based features \cite{asynchronous2007icdar,subtractive2015singh}, descriptor features \cite{histogram2016dwivedi,khan2019dgmm,lai2020pathlet}, \emph{etc}. In the second stage, researchers have successively investigated the Gaussian Mixture Model-Universal Background Model (GMM-UBM) \cite{gmm2008pr}, IR-based techniques (\emph{e.g.}, Term Frequency and Inverse Document Frequency (TF-IDF) \cite{tan2009allograph,tan2010individuality}), and codebook-based methods \cite{subtractive2015singh,histogram2016dwivedi,sparsecoding2018venugopal,lai2020pathlet}.

\subsubsection{Deep Neural Network-Based Methods}
With the prevalence of deep learning, neural networks have emerged as the cutting-edge approaches, with modeling schemes evolving from CNN/RNN-based models \cite{cnn2015fiel,xing2016deepwriter,deepwriterid2016weixin,pslstm2017liu,rnn2017zhang,deepadaptive2019he,fragnet2020he,grrnn2021he} to attention-based models \cite{chen2021level,rstc2022zhang,writeridvit2022koepf}. Fiel and Sablating \cite{cnn2015fiel}, and Yang et al. \cite{deepwriterid2016weixin} first introduced the CNN to offline and online Writer-ID, respectively. Xing et al. \cite{xing2016deepwriter} and He and Schomaker \cite{deepadaptive2019he,fragnet2020he} proposed two-stream CNN networks with different input strategies and feature fusion methods. Liu et al. \cite{pslstm2017liu} and Zhang et al. \cite{rnn2017zhang} both utilized LSTM for online Writer-ID. Moving beyond CNN/RNN, Chen et al. \cite{chen2021level} pioneered in combining CNN, LSTM, and attention mechanism \cite{attention2017vaswani} with various attention-based modules. Recently, the Vision Transformer \cite{dosovitskiy2021an} has become a promising alternative to conventional CNN/RNN by the effective incorporation of the self-attention mechanism. Zhang \cite{rstc2022zhang} exploited the Swin Transformer \cite{swin2021liu} to improve the local fine-grained handwritten feature modeling. Koepf et al. \cite{writeridvit2022koepf} used SIFT to detect handwriting key points and extract image patches, while utilizing ViT-Lite \cite{vitlite2021hassani} for feature encoding.

There is a noticeable trend regarding the granularity of the utilized handwritten data in this field. Over the last two decades, methods were mainly evaluated on page-level or paragraph-level data. For example, Li and Tan \cite{shapecodes2009li} conducted experiments on page-level online document data. Schlapbach et al. \cite{gmm2008pr} and Tan et al. \cite{tan2010individuality} evaluated their methods with handwritten paragraphs. With time evolving, more fine-grained handwritten data has been utilized for Writer-ID, such as handwritten lines or words. It meanwhile endows higher challenges due to limited handwriting features. For instance, Venugopal and Sundaram \cite{sparsecoding2018venugopal} conducted experiments with paragraph and line scripts. Chinese character-level data was analyzed in \cite{deepwriterid2016weixin,pslstm2017liu}. English word-level data was analyzed in \cite{deepadaptive2019he,fragnet2020he,grrnn2021he,rstc2022zhang}. Chen et al. \cite{chen2021level} further used online handwritten letters to perform letter-level Writer-ID, reaching the most fine-grained data level. This trend reflects a growing community focus on addressing practical challenges in Writer-ID, specifically by prioritizing the more difficult yet common identification through words or phrases rather than entire pages. This resonates with our vision that available online handwriting samples in practice are usually short, emphasizing the high applicability of our proposed OLIWER dataset.

\begin{figure*}[t]
	\centering
	\includegraphics[width=\textwidth]{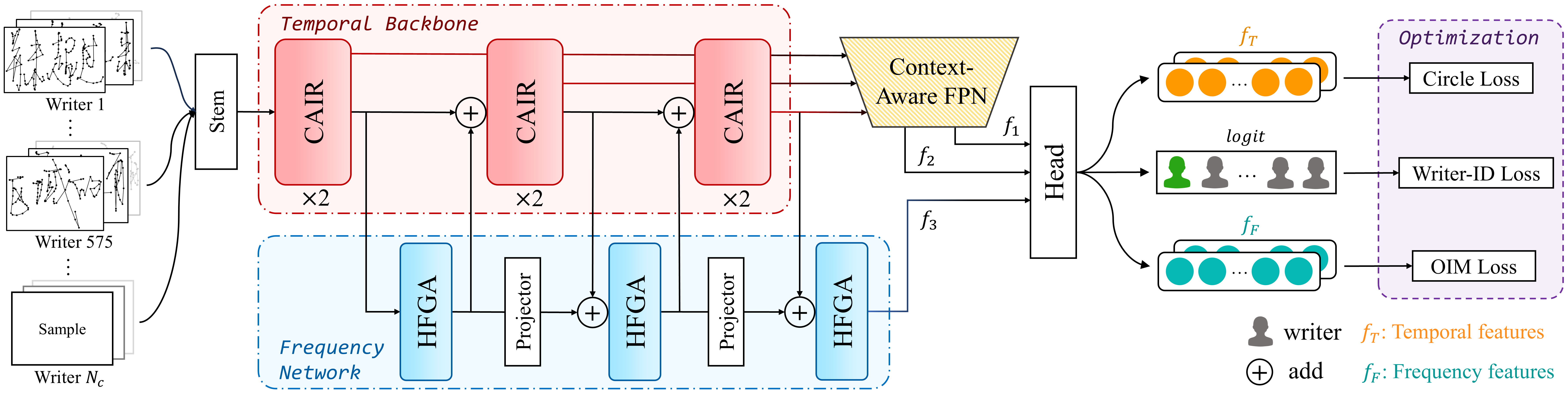}{}
	\caption{The overall architecture of the proposed writer retrieval model DOLPHIN. DOLPHIN is a 1D CNN-based model, which consists of the Temporal Backbone, the Frequency Network, and the Context-Aware FPN. Both \emph{Stem} and \emph{Projector} are depthwise separable convolutions \cite{howard2017mobilenets}, respectively responsible for initial feature projection and internal feature shape matching. \emph{Head} consists of three pooling layers and a multi-layer perceptron to generate final feature embeddings. It outputs temporal feature vectors $f_T$, frequency feature vectors $f_F$, and classification logits for loss computation.}
	\label{Fig::overall}
\end{figure*}

\section{Methodology}
\label{sec::method}
We illustrate the architecture of the writer retrieval model DOLPHIN in Fig.~\ref{Fig::overall}. We begin with inputting online handwriting to \emph{Stem}, a depthwise separable convolution layer \cite{howard2017mobilenets} with a kernel size of 7, to project the time series to high-dimension feature sequences. Subsequently, these sequences undergo synergistic time-frequency analyses by the Temporal Backbone and the Frequency Network. Features extracted by the Temporal Backbone are then fed into the Context-Aware FPN for contextual modeling. Finally, the \emph{Head} block is responsible for pooling the feature sequences from the Context-Aware FPN and the Frequency Network for loss computation.

\begin{figure}[t]
	\centering
	\includegraphics[width=0.78\linewidth]{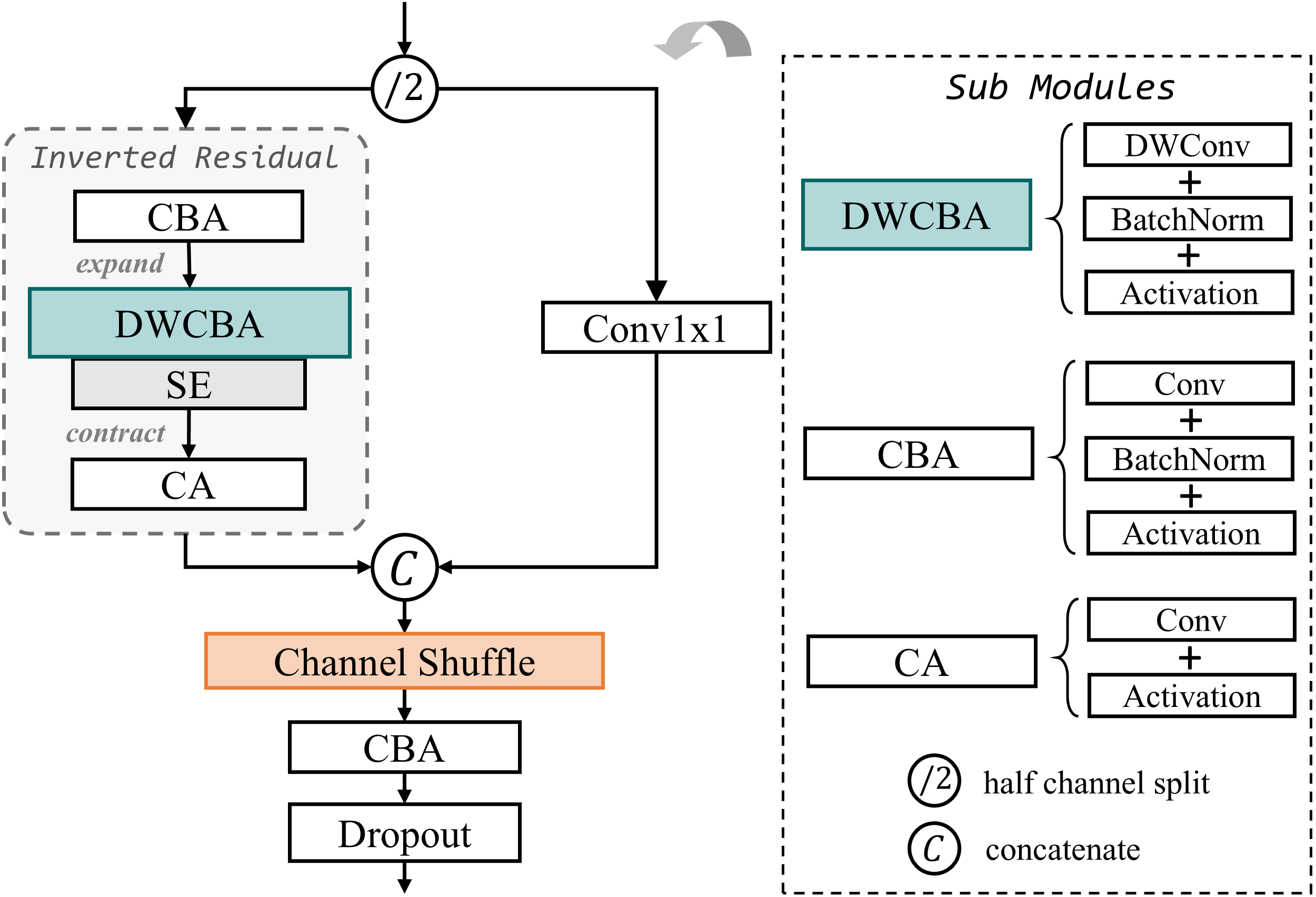}{}
	\caption{Structure of the CAIR block.}
	\label{Fig::cair}
\end{figure}

\subsection{Channel Activation Inverted Residual}
In sequential modeling, projecting channels into a high-dimensional space is a typical practice to facilitate feature learning. However, information redundancy in the channel dimension greatly diminishes the representation capacity for useful knowledge. To reduce channel redundancy, we propose the Channel Activation Inverted Residual (CAIR) block to perform progressive channel activation, as outlined in Fig.~\ref{Fig::cair}. Given an online handwritten sample $x \in \mathbb{R}^{L \times d}$ ($L$ is the sequence length of this sample, $d$ is the channel dimension) as input, it is firstly split in half over the channel dimension to obtain $x_1$ and $x_2$. $x_1$ undergoes processing by an Inverted Residual (IR) \cite{mobilev22018sandler} module, which employs a channel-wise expansion-contraction scheme to extract features. We modify the 2D IR into the 1D version for sequential modeling and add an SE module \cite{senet2018hu} after the \emph{DWCBA} module to enhance channel relationship description, as shown in Fig.~\ref{Fig::cair}. Concurrently, $x_2$ is passed through a $1\times 1$ convolution to identically match the shape of $x_1$ output by the 1D-IR. Then they are concatenated along the channel dimension to yield $x'$, which is further processed by the channel shuffle operation \cite{shufflenet2018zhang} to excite informative channel features. Finally, we pass $x'$ through a \emph{CBA} module and a dropout operation to get the output features. 

The channel split-concat scheme, 1D IR, SE module, and channel shuffle operation progressively activate the information flow across the channel dimension, effectively diminishing the redundancy of channel-wise features and thus empowering channel modeling. For the activation function used in the \emph{CBA} and \emph{CA} modules, we adopt ReLU \cite{relu2010hinton} in the \emph{DWCBA} module and the first \emph{CBA} module, while using SELU \cite{selu2017nips} in all other modules. We stack CAIR blocks to build the Temporal Backbone as illustrated in Fig.~\ref{Fig::overall}, serving as the basic network to learn temporal features of handwriting and interacting with the Frequency Network subsequently.

\subsection{High Frequency Gated Attention}
\begin{figure}[t]
	\centering
	\includegraphics[width=0.65\linewidth]{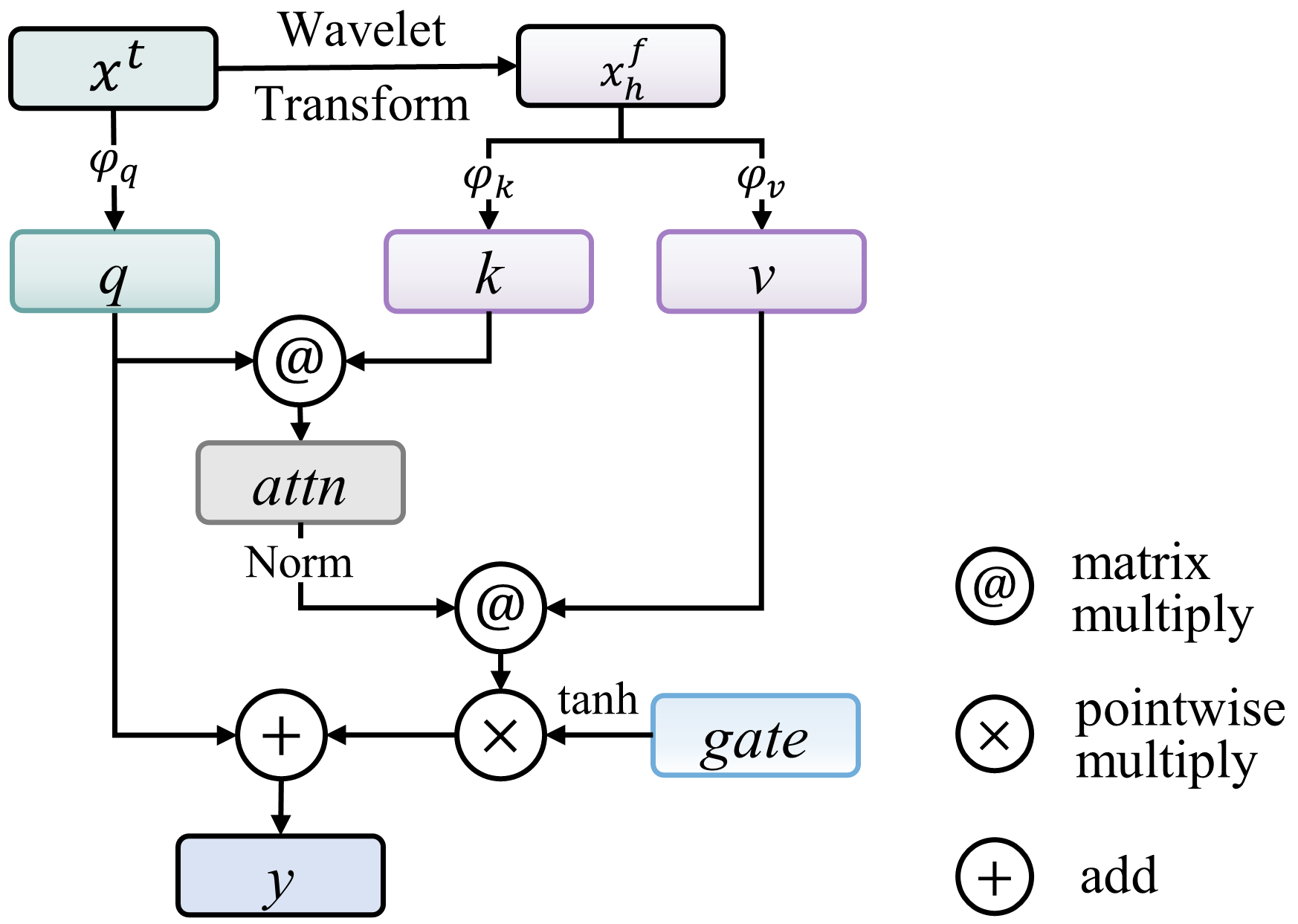}{}
	\caption{Schematic of the HFGA block.}
	\label{Fig::hfga}
\end{figure}

Frequency analysis excels at revealing discriminative writing features that usually remain hidden in the time domain, such as rhythmic patterns and scale-invariant stroke formations that characterize an individual's unique handwriting style. Nevertheless, frequency features have been largely overlooked in online handwriting modeling. To this end, we propose the High Frequency Gated Attention (HFGA) block to capitalize on the rich and distinctive writing features in the high-frequency spectrum. Fig.~\ref{Fig::hfga} depicts the schematic of the HFGA block. Formally, considered a handwritten sample in the temporal domain $x^t \in \mathbb{R}^{L \times d}$, we utilize the 1D Discrete Wavelet Transform (DWT) to decompose $x^t$ into the high frequency components $x^f_h \in \mathbb{R}^{{\lfloor L / 2 \rfloor}\times d}$ and low frequency components $x^f_l \in \mathbb{R}^{{\lfloor L / 2 \rfloor}\times d}$. Here, $x^f_h$ corresponds to the sharp turns and fine-grained details of the signal, which represent local features such as stroke curvatures, pressure variations, and writing speed fluctuations. $\lfloor \cdot \rfloor$ denotes the flooring operation.

\begin{figure}[t]
	\centering
	\includegraphics[width=0.75\linewidth]{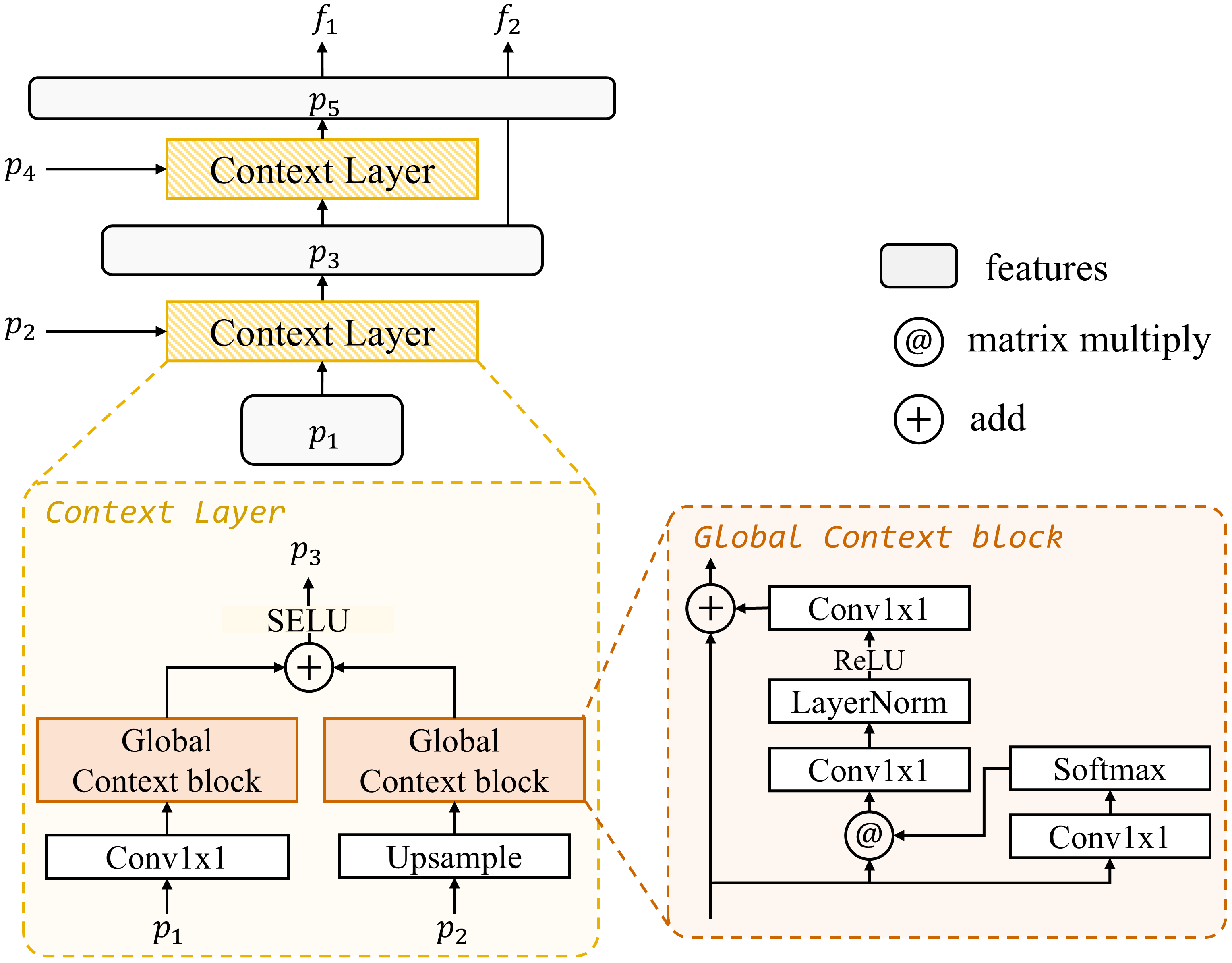}{}
	\caption{Schematic of the Context-Aware FPN.}
	\label{Fig::cfpn}
\end{figure}

Subsequently, we perform gated cross-attention between the vanilla temporal sequence $x$ and high-frequency sub-band $x^f_h$. We transform $x^t$ into query $q \in \mathbb{R}^{L \times \hat{d}}$, and transform $x^f_h$ into $k \in \mathbb{R}^{{\lfloor L / 2 \rfloor} \times \hat{d}}$ and $v \in \mathbb{R}^{{\lfloor L / 2 \rfloor} \times \hat{d}}$, formulated as:
\begin{equation}
	\label{eq1}
	\centering
	\begin{aligned}
	q = \phi_q(x^t); k = \phi_k(x^f_h); v = \phi_v(x^f_h),\\
	\end{aligned}
\end{equation}
where $\phi$ is a $1\times1$ convolution that performs linear transformation. $\hat{d}$ signifies the channel dimension after transformation. We then compute the attention as:
\begin{equation}
	\label{eq2}
	\centering
	w = \frac{k^T\textrm{@}q}{\sqrt{L}}, w \in \mathbb{R}^{L \times {\lfloor L / 2 \rfloor}},
\end{equation}
\begin{equation}
	\label{eq3}
	\centering
	\hat{w} = w - max(w),
\end{equation}
\begin{equation}
	\label{eq4}
	\centering
	y_a = softmax(\hat{w})\textrm{@}v,
\end{equation}
where @ denotes matrix multiplication, $T$ denotes transpose operation. Specifically, $max(w)$ in Eq.~\ref{eq3} represents the maximum value of $w$ along the second dimension of $\mathbb{R}^{\lfloor L / 2 \rfloor}$. This subtraction improves the numerical stability during computation. We refer to Flamingo \cite{flamingo2022nips} and add a gate constraint $g$ to the cross-attention output $y_a$:
\begin{equation}
	\label{eq5}
	\centering
	y = tanh(g) \times y_a + x,
\end{equation}
where $\times$ indicates pointwise multiplication and $g$ is a learnable parameter. We leverage $tanh(\cdot)$ as the gated mechanism to regulate the information flow, and perform a residual connection \cite{resnet2016he} with the original input $x$. The gated cross-attention mechanism enables us to leverage high-frequency components effectively, extracting stylistic writing traits embedded within the frequency features. Our design further achieves the synergy between temporal and frequency analysis, enhancing holistic writing characteristic representation. 

As shown in Fig.~\ref{Fig::overall}, we construct the Frequency Network using three HFGA blocks and two \emph{Projector}s (depthwise separable convolution \cite{howard2017mobilenets}), where the \emph{Projector} aims to match the shape of internal variables for subsequent processing. We further enable the interaction between the Frequency Network and the Temporal Backbone, prompting reciprocal knowledge exchange and feature complementarity. This dynamic interaction not only deepens the joint learning of temporal-frequency features but also empowers the model to capture intricate handwriting patterns across both domains.

\subsection{Context-Aware FPN}
\label{sec::cfpn}
We introduce the Context-Aware Feature Pyramid Network (Context-Aware FPN) for low-level knowledge reusage and multi-scale feature fusion. FPN \cite{fpn2017lin} is initially designed for detecting objects at different scales in the realm of object detection. We directly borrow this manner to reuse the low-level features extracted by the previous layers. To enhance context understanding, we integrate the Global Context (GC) block from GCNet \cite{gcnet2019iccvw} into FPN and introduce the Context-Aware FPN. As illustrated in Fig.~\ref{Fig::cfpn}, it is a two-layer FPN stacked with two Context Layers. We use the bottom Context Layer as an example to elucidate the computation details. Considered the bottom feature sequence $p_1 \in \mathbb{R}^{L_1 \times d_1}$ and the intermediate feature from the upper layer $p_2 \in \mathbb{R}^{L_2 \times d_2}$ as input, where $L_1, L_2$ are the sequence lengths adhering to $L_2 = 2L_1$ and $d_1, d_2$ are the channel dimensions, we derive $p_3$ through a Context Layer as:
\begin{equation}
	\centering
	p'_1 = upsample(p_1), p'_1 \in \mathbb{R}^{L_2 \times d_1},
\end{equation}
\begin{equation}
	\centering
	p'_2 = conv1\times1(p_2), p'_2 \in \mathbb{R}^{L_2 \times d_1},
\end{equation}
\begin{equation}
	\centering
	\begin{aligned}
		&\hat{p_1} = GC\_block_1(p'_1), \hat{p_1} \in \mathbb{R}^{L_2 \times d_1};\\
		&\hat{p_2} = GC\_block_2(p'_2), \hat{p_2} \in \mathbb{R}^{L_2 \times d_1},
	\end{aligned}
\end{equation}
\begin{equation}
	\centering
	p_3 = SELU(\hat{p_1} + \hat{p_2}), p_3 \in \mathbb{R}^{L_2 \times d_1}.
\end{equation}
We upsample $p_1$ using bilinear interpolation to double its sequence length to $L_2$ and get $p'_1$. A $1\times1$ convolution is then utilized to match the channel dimension of $p_2$ to $d_1$, resulting in $p'_2$. Consequently, $p'_1$ and $p'_2$ has the same shape of $\mathbb{R}^{L_2\times d_1}$. The GC block is employed to model the context dependency of the sequence and will maintain its shape unchanged. We add $\hat{p_1}$ and $\hat{p_2}$, and activate the summation with a SELU to get the output $p_3 \in \mathbb{R}^{L_2 \times d_1}$. Similarly, the top Context Layer outputs $p_5 \in \mathbb{R}^{L_4 \times d_1}$, where $L_4$ is the length of sequence $p_4$. We denote $p_5$ and $p_3$ as $f_1$ and $f_2$ for further processing.

The \emph{Head} shown in Fig.~\ref{Fig::overall} resides at the tail of DOLPHIN, comprising three pooling layers, a multi-layer perceptron, and a dropout layer. It takes $f_1$ and $f_2$ outputted by Context-Aware FPN, as well as $f_3$ yielded by Frequency Network as input. Each of them is transformed into fixed-length feature vectors through a pooling layer as:
\begin{equation}
	\centering
	\begin{aligned}
		f_1^{'}=pool(f_1); f_2^{'}=pool(f_2); f_3^{'}=pool(f_3).\\
	\end{aligned}
\end{equation}
We adopt the Selective Pooling \cite{lai2021synsig2vec} as the pooling scheme. This results in $f_1 \in \mathbb{R}^{L_4 \times d_1} \rightarrow f_1^{'} \in \mathbb{R}^{d_1}$, $f_2 \in \mathbb{R}^{L_2 \times d_1} \rightarrow f_2^{'} \in \mathbb{R}^{d_1}$, and $f_3 \in \mathbb{R}^{L_F \times d_1} \rightarrow f_3^{'} \in \mathbb{R}^{d_1}$, where $L_4$ is the length of sequence $p_4$ and $L_F$ is the length of $f_3$.

Subsequently, we obtain the final temporal feature embeddings $f_T$, classification logits $logit$, and final frequency feature embeddings $f_F$ through:
\begin{equation}
	\centering
	\label{eq::fT}
	f_T = dropout(W^T_1 \cdot concat(f_1,f_2)), W_1 \in \mathbb{R}^{2d_1 \times d_1},
\end{equation}
\begin{equation}
	\centering
	logit = W_l^T \cdot f_T, W_l \in \mathbb{R}^{d_1 \times N_c},
\end{equation}
\begin{equation}
	\label{eq11}
	\centering
	f_F = dropout(W^T_3 \cdot f_3), W_3 \in \mathbb{R}^{d_1 \times d_1},
\end{equation}
where $concat(\cdot)$ denotes the concatenation operation, $N_c$ is the number of writers. $W_l^T$ is the weight of the final classification layer. $f_T$, $logit$, and $f_F$ are fed into loss computation for model optimization.

\subsection{Model Optimization}
\label{sec::optimization}
During the retrieval phase, we use DOLPHIN to extract individualistic feature embeddings for input handwriting and perform similarity-based matching, which will be detailed in the next subsection. Therefore, we exploit several metric-learning loss functions to conduct vector-space optimization. We adopt the Circle Loss \cite{circleloss2020sun} for optimizing $f_T$ and the Online Instance Matching (OIM) Loss \cite{oimloss2017xiao} for optimizing $f_F$. The Circle Loss is formulated as:
\begin{equation}
	\centering
	\mathcal{L}_{circle} = log[1 + \sum_{j=1}^{|\mathcal{P}|} e^{\gamma \alpha^j_n(s_n^j - \delta_n)}\sum_{i=1}^{|\mathcal{N}|}e^{-\gamma \alpha_p^i(s_p^i - \delta_p)}],
\end{equation}
\begin{equation}
	\centering
	\begin{aligned}
		&\alpha_p^i = [O_p - s_p^i]_+;\\
		&\alpha_n^j = [s_n^j - O_n]_+,
	\end{aligned}
\end{equation}
\begin{equation}
	\centering
	\label{eq15}
	\begin{aligned}
		&O_p = 1 + m; O_n = -m;\\
		&\delta_p = 1 - m; \delta_n = m,
	\end{aligned}
\end{equation}
\noindent where $s_p$ denotes cosine similarities of samples from the same writer, whereas $s_n$ denotes cosine similarities of samples from different writers. $|\mathcal{P}|$ is the number of positive samples and $|\mathcal{N}|$ is the number of negative samples. $O_p$ and $O_n$ are respectively the optimal values of $s_p^i$ and $s_n^j$. Sun et al. \cite{circleloss2020sun} simplify the hyper-parameters by setting a relaxation margin $m$ as in Eq.~\ref{eq15}. $\gamma$ is a scale factor. 

The OIM Loss is formulated as:
\begin{equation}
	\centering
	\begin{aligned}
	&p_i = softmax(s_i \textrm{@} V^T), V \in \mathbb{R}^{N_c \times d_1},\\
	&\mathcal{L}_{oim} = E_x[log(p_i)],i \in \{1,2,...,N_c\},
	\end{aligned}
\end{equation}
\noindent where $s_i$ is a feature vector of $\mathbb{R}^{d_1}$, $N_c$ is the number of writers, and $V$ is the weights of a lookup table. The $d_1$ and $N_c$ align with the definitions in Sec.~\ref{sec::cfpn}.

In addition, open-set retrieval requires the model to distinguish handwriting of different writers, even though the writers are unseen during training. Therefore, we leverage the training data writers as surrogate classes to cultivate this ability. We adopt the cross-entropy loss with label smoothing $\epsilon$ as the Writer-ID loss $L_{ID}$, receiving the multi-classification logits $logit$ for optimization. This supervision allows the model to better differentiate unseen testing writers.

Eventually, the full optimization objective is:
\begin{equation}
	\centering
	\mathcal{L} = \mathcal{L}_{circle} + \mathcal{L}_{oim} + \mathcal{L}_{ID}.
\end{equation}

\subsection{Retrieval Procedure}
\label{sec::ret_procedure}
To retrieve the handwritten samples of the specific writer, we leverage the feature representation outputted by the model for comparison and matching. Given a query handwriting and an existing database, \emph{a.k.a.} gallery, we use the model to extract feature embeddings of query and all samples in the gallery. Then, we compute one-to-one cosine similarity between query embedding and each gallery sample embedding, where higher similarities indicate a larger likelihood of the two handwritten samples originating from the same individual. By ranking the gallery samples based on the similarities, we can retrieve the most similar handwriting as potential matches. For experiments with DOLPHIN, we utilize the feature embedding $f_T$ (Eq.~\ref{eq::fT}) as the feature representation of an input handwriting. For other models, we use the feature embedding outputted before the last classification layer as the feature representation.

\section{OLIWER Dataset}
\label{sec::create_dataset}

\begin{table}[t]
	\renewcommand{\arraystretch}{1.2}
	\caption{Details of the OLIWER dataset and the three constituent datasets. \emph{Org.} denotes original and \emph{Seg.} represents segmented. \emph{P.S.F} denotes the point sampling frequency during raw data collection.}
	\label{Table::datasets}
	\centering
	\resizebox{\linewidth}{!}{
		\begin{tabular}{c c c c c c}
			\whline{1.1pt}
			\textbf{Dataset} & \textbf{\#Writer} & \textbf{\#Org. Sample} & \textbf{\#Seg. Sample} & \textbf{P. S. F.} & \textbf{Features}\\
			\hline
			CASIA-OLHWDB \cite{casiaolhwdb2011liu} & 1,019 & 52,220 & 330,469 & 30Hz & $X$,$Y$\\
			\hline
			DCOH-E \cite{DCOH2024dlvc} & 567 & 87,759 & 198,548 & 120Hz & $X$,$Y$,$P$,$T$\\
			\hline
			SCUT-COUCH2009 \cite{couch20092011jin} & 145 & 145,000 & 145,000 & 30Hz & $X$,$Y$\\
			\hline
			OLIWER & 1,731 & - & 674,017 & - & $X$,$Y$,$P$\\
			\whline{1.1pt}
	\end{tabular}}
\end{table}
\unskip

The OLIWER dataset is an online writer retrieval dataset restructured from three public online handwriting datasets, namely CASIA-OLHWDB \cite{casiaolhwdb2011liu}, DCOH \cite{DCOH2024dlvc}, and SCUT-COUCH2009 \cite{couch20092011jin}, in which we extend the DCOH dataset by incorporating data from additional writers. The construction details are elaborated below.

An online handwriting sample $S$ is typically represented as a time series consisting of discrete points $S=\{ p_1,p_2,...,p_N \}$, where $N$ is the number of points. Each point $p_i, i \in \{1,..., N\}$ contains the dynamic information captured during the writing process, including $x$ and $y$ coordinates, pressure $p$, \emph{etc.} Since these specifics may vary in distinct datasets, we standardize the format to ensure that each sample uniformly includes at least three types of dynamic information: $x$, $y$, and $p$. Surplus information provided in the original dataset is preserved and the missing attributes are supplemented manually. Subsequently, we conduct specific processing on different datasets to acquire consecutively handwritten phrases within two to five characters:
\begin{itemize}
	\item \textbf{CASIA-OLHWDB.} CASIA-OLHWDB includes CASIA-OLHWDB1.0-1.2 subsets for isolated Chinese characters and CASIA-OLHWDB2.0-2.2 subsets for Chinese text lines. Instead of splicing isolated characters, which might disrupt the continuity of handwriting features, we segment 52,220 handwritten lines from 1,019 writers of the CASIA-OLHWDB2.0-2.2 subsets to acquire consecutively written phrases. Since the character annotations, \emph{i.e.}, the character belonging to each point, are provided, we directly combine the well-segmented characters to generate phrases, in which the character numbers are randomly selected between two to five. This process preserves the coherence and naturalness of the handwritten phrases. As merely $x$ and $y$ are available, we add the pressure information $p=1$ to each point.
	
	\item \textbf{DCOH-E.} We use the Chinese subset of the DCOH dataset, containing 82,659 handwritten lines from 313 writers, and extend it by adding 5,100 lines from 255 additional writers, resulting in the DCOH-E (E for extended) dataset with 87,759 lines from 567 writers. This dataset provides comprehensive $x,y$ coordinates, pressure $p$, and timestamp $t$ for each point. Although it lacks character-level annotations, stroke-level annotations are available. Therefore, we calculate time intervals between strokes according to timestamps and divide each line into separate characters based on these intervals (details are included in Procedure~\ref{algorithm::seg} in supplementary files). With the processed isolated characters, we combine them to form phrases. Notably, the segmentation algorithm avoids compromising the natural flow of the handwriting, which well maintains the writing integrity and consecutiveness.
	
	\item \textbf{SCUT-COUCH2009.} This dataset contains frequently used Chinese words/phrases and consists of three subsets. These subsets are contributed by 130, 10, and 5 writers, with 8,888, 17,366, and 44,208 handwritten samples per user, respectively. We randomly select 1,000 words for each writer and obtain a total of 145,000 samples from 145 writers. Since only $x,y$ coordinates of the online handwriting are provided, we manually add the pressure $p=1$ to each point.
\end{itemize}

\begin{algorithm}[t]
	\caption{Procedure for segmenting online handwritten lines into online handwritten phrases, with only stroke-level annotations of the data.}
	\label{algorithm::seg}
	\begin{algorithmic}
		\State \begin{itemize}
			\item[(1)] Given a text line $X$ composed of a set of strokes $X$=$\{s_1,s_2,...,s_N\}$, where $s_i,i\in N$ denotes a stroke composed of multiple points and $N$ is the total number of strokes.
			\item[(2)] Since timestamps are pre-known, we can calculate the time interval $ts_{i-1}$ between stroke $s_{i-1}$ and stroke $s_i$ by subtracting the first timestamp of $s_i$ from the last timestamp of $s_{i-1}$, and get a sequence of time intervals $T=\{ ts_1,ts_2,...,ts_{N-1}\}$.
			\item[(3)] Given the text transcription $Y$ of line $X$, we calculate the number of Chinese characters in the transcription, denoting as $N_Y$. We sort $T$ in descending order and select the $ts$ ranked in $N_Y + 1$ position as a time threshold $d$. In this case, there should be $N_Y$ time intervals $\textgreater~d$ and we denote them as $T^d = \{ ts_1^d,ts_2^d,...,ts_{N_Y}^d\}$, while the rest are $\le d$.
			\item[(4)] We sequentially traverse all the strokes in $X$ and utilize the time intervals in $T^d$ as division points to cluster the strokes between two division points, in which each cluster represents an isolated character. This leads to in total $N_Y$ isolated characters segmented from the given text line $X$.
		\end{itemize}
	\end{algorithmic}
\end{algorithm}

Ultimately, we consolidate the consecutively handwritten phrases from the three datasets to create the OLIWER dataset, excluding writers with fewer than 20 samples. This results in a final dataset of 674,017 online handwritten samples from 1,731 writers. The details of the three component datasets and the OLIWER dataset are summarized in Table~\ref{Table::datasets}.

\section{Experiment}
\label{sec::experiment}
\subsection{Dataset}
\label{sec::exp_dataset}
We conduct experiments across four datasets: the newly introduced OLIWER dataset, the segmented CASIA-OLHWDB2 dataset \cite{casiaolhwdb2011liu}, the segmented DCOH-E dataset \cite{DCOH2024dlvc}, and the SCUT-COUCH2009 dataset \cite{couch20092011jin} that undergoes sample selection (Sec.~\ref{sec::create_dataset}). For fair comparisons, we employ a consistent data-splitting strategy across all datasets, assigning the samples of 80\% of the writers as the training set and leaving the remaining 20\% for testing. This achieves an open-set setting, where the testing writers are entirely unseen during training. Following common practices in other retrieval-related domains like person re-identification \cite{osnet2022zhou,cal2021rao,cdnet2021li}, we split the test data into a query set and a gallery set to simulate real-world retrieval scenarios. The query set is composed of one randomly selected sample from each writer, while the gallery set consists of the remaining samples from all writers. This results in 538,209/347/135,461 training/query/gallery samples from 1,384/347/347 writers for OLIWER, 264,161/204/66,104 samples from 815/204/204 writers for CASIA-OLHWDB2, 163,986/114/34,448 samples from 453/114/114 writers for DCOH-E, and 116,000/29/28,971 samples from 116/29/29 writers for SCUT-COUCH2009.

\begin{table}[t]
	\renewcommand{\arraystretch}{1.1}
	\caption{Time functions.}
	\label{Table::timefunc}
	\centering
	\resizebox{0.73\linewidth}{!}{
		\begin{tabular}{c c}
			\whline{1.1pt}
			\textbf{\#} & \textbf{Features}\\
			\hline
			1 & First-order derivative of Coordinate $x$: $\dot{x}$\\
			\hline
			2 & First-order derivative of Coordinate $y$: $\dot{y}$\\
			\hline
			3 & Second-order derivative of Coordinate $x$: $\ddot{x}$\\
			\hline
			4 & Second-order derivative of Coordinate $y$: $\ddot{y}$\\
			\hline
			5 & Velocity magnitude: $v = \sqrt{\dot{x}^2 + \dot{y}^2}$\\
			\hline
			6 & Path-tangent angle: $\theta = \arctan{\frac{\dot{y}}{\dot{x}}}$\\
			\hline
			7 & Cosine of the path-tangent angle: $\cos{\theta}$\\
			\hline
			8 & Sine of the path-tangent angle: $\sin{\theta}$\\
			\hline
			9 & First-order derivative of $v$: $\dot{v}$\\
			\hline
			10 & First-order derivative of $\theta$: $\dot{\theta}$\\
			\hline
			11 & Log curvature radius: $\rho = \log{\frac{v}{\dot{\theta}}}$\\
			\hline
			12 & Centripetal acceleration magnitude: $\bigtriangleup v =  v \cdot \dot{\theta}$\\
			\hline
			13 & Total acceleration magnitude: $a = \sqrt{\dot{v}^2 + \bigtriangleup v^2}$\\
			\hline
			14 & Pressure: $p$\\
			\whline{1.1pt}
	\end{tabular}}
\end{table}

\begin{table*}[t]
	\renewcommand{\arraystretch}{1.1}
	\caption{Comparison between DOLPHIN and other State-of-the-Art methods on the OLIWER dataset, the segmented CASIA-OLHWDB2 dataset, the segmented DCOH-E dataset, and the SCUT-COUCH2009 dataset. \emph{On-SV} denotes online signature verification. \emph{Off-WID} denotes offline writer identification. \emph{Re-ID} denotes person re-identification. \emph{Off-WR} denotes offline writer retrieval. \emph{On-WR} denotes online writer retrieval. $\uparrow$ signifies that the higher value is better. Each experiment is repeated 50 times. Results are reported as $avg$ (\textpm $std$), where $avg$ is the average performance and $std$ is the standard deviation. The best performances are marked in \textbf{bold} and the second-best results are marked with \underline{underline}.}
	\label{Table::sota-comp}
	\begin{minipage}{\textwidth}
		\centering
		\resizebox{\linewidth}{!}{
			\begin{tabular}{c c c c c c c | c c c c}
				\whline{1.05pt}
				\noalign{\vspace{3pt}}
				\multirow{2.3}{*}{\textbf{Model}} &
				\multirow{2.3}{*}{\textbf{Venue}} & \multirow{2.3}{*}{\textbf{Domain}} & \multicolumn{4}{c}{\textbf{OLIWER}} & \multicolumn{4}{c}{\textbf{CASIA-OLHWDB2}}\\
				\cmidrule(r){4-7}\cmidrule(r){8-11}
				~ & ~ & ~ & \textbf{R1} $\uparrow$ & \textbf{R5} $\uparrow$ & \textbf{R10} $\uparrow$ & \textbf{mAP} $\uparrow$ & \textbf{R1} $\uparrow$ & \textbf{R5} $\uparrow$ & \textbf{R10} $\uparrow$ & \textbf{mAP} $\uparrow$\\
				\hline
				ResNet-34 \cite{resnet2016he} & CVPR'16 & General & 84.43 (\textpm 1.90) & 96.08 (\textpm 1.09) & 97.90 (\textpm 0.79) & 47.86 (\textpm 1.05) & 85.50 (\textpm 2.01) & 95.62 (\textpm 1.32) & 97.56 (\textpm 1.01) & 47.08 (\textpm 1.30)\\
				MobileNetV2 \cite{mobilev22018sandler} & CVPR'18 & General & 83.13 (\textpm 1.83) & 95.42 (\textpm 1.21) & 97.54 (\textpm 0.92) & 46.69 (\textpm 0.97) & 83.60 (\textpm 2.78) & 95.30 (\textpm 1.64) & 97.34 (\textpm 1.27) & 45.00 (\textpm 1.37)\\
				EfficientNet-b0 \cite{effnet2019tan} & ICML'19 & General & 82.28 (\textpm 1.96) & 94.79 (\textpm 1.14) & 97.20 (\textpm 0.86) & 41.35 (\textpm 0.97) & 86.42 (\textpm 2.34) & 95.92 (\textpm 1.52) & 97.51 (\textpm 1.12) & 46.04 (\textpm 1.37)\\
				ConvNeXt-Tiny \cite{convnext2022liu} & CVPR'22 & General & 76.03 (\textpm 2.23) & 91.19 (\textpm 1.41) & 94.72 (\textpm 1.09) & 38.03 (\textpm 0.96) & 75.90 (\textpm 3.24) & 90.81 (\textpm 2.38) & 94.39 (\textpm 2.00) & 35.12 (\textpm 1.10)\\
				HorNet-Tiny \cite{hornet2022rao} & NeurIPS'22 & General & 77.96 (\textpm 1.86) & 92.69 (\textpm 1.54) & 95.90 (\textpm 1.11) & 40.26 (\textpm 1.00) & 84.03 (\textpm 2.39) & 95.19 (\textpm 1.38) & 97.08 (\textpm 1.06) & 43.46 (\textpm 1.25)\\
				FasterNet \cite{fasternet2023chen} & CVPR'23 & General & 82.48 (\textpm 2.30) & 94.82 (\textpm 1.17) & 97.04 (\textpm 0.94) & 45.11 (\textpm 0.94) & 71.96 (\textpm 3.33) & 89.78 (\textpm 2.14) & 93.94 (\textpm 1.89) & 34.82 (\textpm 1.26)\\
				GR-RNN \cite{grrnn2021he} & PR'21 & Off-WID & 89.95 (\textpm 1.52) & 97.50 (\textpm 0.89) & 98.73 (\textpm 0.69) & 52.68 (\textpm 1.05) & 89.95 (\textpm 2.09) & 97.34 (\textpm 1.31) & 98.53 (\textpm 0.97) & 53.03 (\textpm 1.60)\\
				Sig2Vec \cite{lai2021synsig2vec} & TPAMI'22 & On-SV & {92.63 (\textpm 1.45)} & {98.24 (\textpm 0.63)} & 99.10 (\textpm 0.50) & 54.70 (\textpm 1.06) & 94.52 (\textpm 1.56) & 98.47 (\textpm 1.04) & {99.06 (\textpm 0.82)} & 59.86 (\textpm 1.57)\\
				CDNet \cite{cdnet2021li} & CVPR'21 & Re-ID & {92.01 (\textpm 1.24)} & {98.12 (\textpm 0.51)} & 99.01 (\textpm 0.47) & 59.73 (\textpm 1.07) & \underline{94.55 (\textpm 1.43)} & \underline{98.53 (\textpm 1.00)} & \underline{99.08 (\textpm 0.76)} & \underline{64.33 (\textpm 1.49)}\\
				CAL \cite{cal2021rao} & ICCV'21 & Re-ID & \underline{93.32 (\textpm 1.23)} & \underline{98.29 (\textpm 0.75)} & \underline{99.16 (\textpm 0.51)} & \underline{61.01 (\textpm 1.10)} & 91.44 (\textpm 2.30) & 97.39 (\textpm 1.19) & 98.32 (\textpm 0.97) & 55.50 (\textpm 1.57)\\
				OSNet \cite{osnet2022zhou} & TPAMI'22 & Re-ID & {91.11 (\textpm 1.64)} & {97.81 (\textpm 0.82)} & {98.92 (\textpm 0.62)} & 57.42 (\textpm 1.10) & {93.09 (\textpm 1.78)} & 98.02 (\textpm 0.98) & {98.78 (\textpm 0.83)} & 61.91 (\textpm 1.52)\\
				NetRVLAD \cite{rethis2023icdar} & ICDAR'23 & Off-WR & 77.02 (\textpm 2.27) & 92.61 (\textpm 1.37) & 95.82 (\textpm 0.92) & 31.48 (\textpm 0.87) & 89.64 (\textpm 2.00) & 96.94 (\textpm 1.30) & 98.07 (\textpm 1.12) & 51.57 (\textpm 1.54)\\
				\hline
				DOLPHIN (\textbf{Ours}) & This work & On-WR & \textbf{96.40 (\textpm 0.94)} & \textbf{99.20 (\textpm 0.48)} & \textbf{99.50 (\textpm 0.37)} & \textbf{69.83 (\textpm 0.37)} & \textbf{97.00 (\textpm 1.23)} & \textbf{99.09 (\textpm 0.69)} & \textbf{99.39 (\textpm 0.58)} & \textbf{71.97 (\textpm 1.47)}\\
				\whline{1.05pt}
		\end{tabular}}
	\end{minipage}
	\vfill
	\vspace{3pt}
	\begin{minipage}{\textwidth}
		\centering
		\resizebox{\linewidth}{!}{
			\begin{tabular}{c c c c c c c | c c c c}
				\whline{1.05pt}
				\noalign{\vspace{3pt}}
				\multirow{2.3}{*}{\textbf{Model}} &
				\multirow{2.3}{*}{\textbf{Venue}} & \multirow{2.3}{*}{\textbf{Domain}} & \multicolumn{4}{c}{\textbf{DCOH-E}} & \multicolumn{4}{c}{\textbf{SCUT-COUCH2009}}\\
				\cmidrule(r){4-7}\cmidrule(r){8-11}
				~ & ~ & ~ & \textbf{R1} $\uparrow$ & \textbf{R5} $\uparrow$ & \textbf{R10} $\uparrow$ & \textbf{mAP} $\uparrow$ & \textbf{R1} $\uparrow$ & \textbf{R5} $\uparrow$ & \textbf{R10} $\uparrow$ & \textbf{mAP} $\uparrow$\\
				\hline
				ResNet-34 \cite{resnet2016he} & CVPR'16 & General & 84.14 (\textpm 3.74) & 96.05 (\textpm 1.85) & 98.18 (\textpm 1.21) & 47.98 (\textpm 1.51) & 65.93 (\textpm 9.45) & 91.31 (\textpm 4.54) & {95.38 (\textpm 3.56)} & 32.90 (\textpm 1.95)\\
				MobileNetV2 \cite{mobilev22018sandler} & CVPR'18 & General & 84.19 (\textpm 3.39) & 95.79 (\textpm 1.77) & 98.07 (\textpm 1.31) & 48.90 (\textpm 1.61) & 71.79 (\textpm 9.01) & 92.48 (\textpm 4.76) & {96.21 (\textpm 3.47)} & 37.12 (\textpm 2.76)\\
				EfficientNet-b0 \cite{effnet2019tan} & ICML'19 & General & 74.77 (\textpm 3.74) & 92.51 (\textpm 2.00) & 96.02 (\textpm 1.33) & 37.39 (\textpm 1.34) & 63.17 (\textpm 7.51) & 88.48 (\textpm 5.23) & 93.59 (\textpm 5.02) & 28.53 (\textpm 2.28)\\
				ConvNeXt-Tiny \cite{convnext2022liu} & CVPR'22 & General & 68.46 (\textpm 4.13) & 87.37 (\textpm 3.10) & 91.42 (\textpm 2.66) & 30.55 (\textpm 1.52) & 39.24 (\textpm 7.95) & 69.52 (\textpm 8.46) & 79.45 (\textpm 7.00) & 17.30 (\textpm 1.83)\\
				HorNet-Tiny \cite{hornet2022rao} & NeurIPS'22 & General & 62.61 (\textpm 3.99) & 85.07 (\textpm 3.21) & 91.05 (\textpm 2.76) & 27.88 (\textpm 1.50) & 37.03 (\textpm 7.55) & 68.48 (\textpm 8.25) & 81.38 (\textpm 5.65) & 15.27 (\textpm 1.66)\\
				FasterNet \cite{fasternet2023chen} & CVPR'23 & General & 81.37 (\textpm 3.73) & 94.77 (\textpm 2.24) & 97.47 (\textpm 1.33) & 45.55 (\textpm 2.03) & 74.69 (\textpm 7.08) & {92.90 (\textpm 4.27)} & {96.14 (\textpm 3.42)} & 41.37 (\textpm 2.52)\\
				GR-RNN \cite{grrnn2021he} & PR'21 & Off-WID & 88.30 (\textpm 2.77) & 97.16 (\textpm 1.64) & 98.47 (\textpm 1.12) & 44.73 (\textpm 1.87) & 69.38 (\textpm 8.18) & 91.45 (\textpm 4.74) & {95.31 (\textpm 3.36)} & 32.48 (\textpm 2.31)\\
				Sig2Vec \cite{lai2021synsig2vec} & TPAMI'22 & On-SV & \underline{94.98 (\textpm 1.84)} & \underline{99.04 (\textpm 0.81)} & \underline{99.65 (\textpm 0.55)} & \underline{60.19 (\textpm 1.43)} & \underline{86.55 (\textpm 5.86)} & \underline{97.47 (\textpm 3.20)} & \underline{99.20 (\textpm 1.71)} & \underline{45.48 (\textpm 2.43)}\\
				CDNet \cite{cdnet2021li} & CVPR'21 & Re-ID & 91.58 (\textpm 2.28) & 98.32 (\textpm 0.99) & 99.02 (\textpm 0.87) & 55.27 (\textpm 1.57) & 82.41 (\textpm 5.37) & 97.01 (\textpm 2.92) & 98.74 (\textpm 1.89) & 44.45 (\textpm 2.68)\\
				CAL \cite{cal2021rao} & ICCV'21 & Re-ID & 91.30 (\textpm 2.56) & 98.07 (\textpm 1.02) & 99.07 (\textpm 0.81) & 54.05 (\textpm 1.54) & 78.62 (\textpm 7.07) & {95.66 (\textpm 3.89)} & 98.07 (\textpm 2.40) & 37.16 (\textpm 2.56)\\
				OSNet \cite{osnet2022zhou} & TPAMI'22 & Re-ID & 91.18 (\textpm 2.19) & 97.82 (\textpm 1.32) & 98.98 (\textpm 0.87) & 55.16 (\textpm 1.42) & 78.48 (\textpm 7.22) & 95.03 (\textpm 4.63) & 98.21 (\textpm 2.69) & 38.13 (\textpm 2.56)\\
				NetRVLAD \cite{rethis2023icdar} & ICDAR'23 & Off-WR & 80.39 (\textpm 3.46) & 93.61 (\textpm 2.25) & 96.04 (\textpm 1.86) & 34.04 (\textpm 1.60) & 70.69 (\textpm 8.54) & 92.34 (\textpm 4.59) & 95.66 (\textpm 3.89) & 34.54 (\textpm 2.53)\\
				\hline
				DOLPHIN (\textbf{Ours}) & This work & On-WR & \textbf{96.97 (\textpm 1.35)} & \textbf{99.53 (\textpm 0.68)} & \textbf{99.81 (\textpm 0.40)} & \textbf{68.23 (\textpm 1.41)} & \textbf{88.16 (\textpm 5.17)} & \textbf{97.82 (\textpm 2.59)} & \textbf{99.43 (\textpm 1.56)} & \textbf{47.68 (\textpm 3.16)}\\
				\whline{1.05pt}
		\end{tabular}}
	\end{minipage}
\end{table*}

\subsection{Evaluation Metric}
\label{sec::metric}
We adopt the Cumulative Matching Characteristic (CMC) curves and the mean Average Precision (mAP) as the evaluation metrics, which are commonly used in retrieval-pertinent literature \cite{rethis2023icdar,tricnn2018keglevic,osnet2022zhou,cdnet2021li,cal2021rao}.

CMC curves include the Rank-1, Rank-5, and Rank-10 accuracies. Given $N_q$ query samples and $N_g$ gallery samples, we can obtain the cosine similarities $s(q_i,g_j), i \in \{1,...,N_q\}, j \in \{1,...,N_g\}$ between each query sample $q_i$ and gallery sample $g_j$. We then rank the similarities in the descending order. For a given ranking $k$, we acquire the top $k$ matching results $\{r_1,r_2,...,r_k\}$ to calculate the matching accuracy $p_k$ as:
\begin{equation}
	\centering
	p_k = \frac{1}{N_q}\sum^{N_q}_{i=1}[q_i \in Top_k(q, \{ r_1,...,r_k \})],
\end{equation}
\noindent where $Top_k(q,\{ r_1,...,r_k \})$ denotes whether the top $k$ gallery matches contain samples from the same writer as $q_i$. With $k$ setting to 1, 5, and 10, we can obtain the corresponding Rank-1, Rank-5, and Rank-10 accuracies. We denote them as R1, R5, and R10, respectively.

mAP is used to evaluate the overall precision across various recall levels of all queries. Assuming that there are $M$ samples belonging to the queried writer $i$ in the gallery set, the Average Precision $AP_i$ is computed as:
\begin{equation}
	\centering
	AP_i = \frac{1}{M}\sum^M_{m=1}\frac{m}{rank(m)},
\end{equation}
\noindent where $rank(m)$ denotes the rank of the $m^{th}$ retrieved sample in the gallery set, obtained by sorting the cosine similarities between the query and all gallery samples. Then we can compute the mean of APs of all queries:
\begin{equation}
	\centering
	mAP = \frac{1}{N_q}\sum^{N_q}_{i=1}AP_i.
\end{equation}
\noindent mAP is considered a more reliable \cite{abdnet2019chen} and challenging metric than CMC curves. It poses a more rigorous criterion as it demands consistent model performance across all retrieval positions, not merely at the top ranks, which better reflects model capabilities. All metrics are reported as percentages.

\subsection{Data Preprocessing}
\label{sec::preprocessing}
We utilize the $x$ coordinates, $y$ coordinates, and pressure $p$ of the raw online handwritten data as the initial features for preprocessing. To mitigate variations in size and location, we perform center normalization on $x$ and $y$, relocating the handwriting center to (0,0) and normalizing coordinate values to the range of (-1,1), while preserving the aspect ratio. A min-max normalization is also applied to the pressure information. Subsequently, we resample all the sequences into 120Hz using bi-cubic interpolation. Furthermore, inspired by the preprocessing in other handwriting analysis domains, such as online signature verification \cite{lai2021synsig2vec,msds2022zhang}, we extract 14 time functions based on the normalized $x$, $y$, and $p$, as outlined in Table~\ref{Table::timefunc}, resulting in an input dimension of 14. All experimental models take the same time functions as input, ensuring fair comparisons. The z-score normalization is applied to the time functions to standardize them with zeros means and unit variance, facilitating model learning.

\subsection{Implementation Detail}
We train DOLPHIN from scratch for 85 epochs with a batch size of 72. AdamW \cite{loshchilov2018decoupled} with $\beta_1=0.9$, $\beta_2=0.999$, and weight decay of 1e-5 is adopted as the optimizer. The learning rate is initially set to 1e-3 and is descended by multiplying a factor of 0.9 after each epoch. The dropout rate in every layer is set to 0.1. $m$ and $\gamma$ in $\mathcal{L}_{circle}$ are respectively set to 0.25 and 32. $\epsilon$ in $\mathcal{L}_{ID}$ is set to 0.1. To maintain uniformity and correctness of implementation, we slightly modify the off-the-shelf code\footnote{\url{https://github.com/layumi/Person_reID_baseline_pytorch}} as our evaluation metric computation code. In experiments with models other than DOLPHIN, we use the circle loss $\mathcal{L}_{circle}$ and the writer-ID loss $\mathcal{L}_{ID}$ to form $\mathcal{L}'=L_{circle} + L_{ID}$ for supervision, where $\mathcal{L}_{circle}$ is to perform metric-learning optimization with the feature embedding computed before the classification layer, and $\mathcal{L}_{ID}$ is utilized to optimize the multi-classification logits, the same as in Sec.~\ref{sec::optimization}.

\subsection{Comparison with State-of-the-Art Method}
We compare our proposed DOLPHIN with the State-of-the-Art (SOTA) methods on the OLIWER dataset, the segmented CASIA-OLHWDB2 dataset \cite{casiaolhwdb2011liu}, the segmented DCOH-E dataset \cite{DCOH2024dlvc}, and the SCUT-COUCH2009 dataset \cite{couch20092011jin}. Due to the scarcity of established methods tailored for online writer retrieval, we draw upon SOTA methods from other domains, including the generic domain \cite{resnet2016he,mobilev22018sandler,effnet2019tan,hornet2022rao,fasternet2023chen,convnext2022liu}, writer identification \cite{grrnn2021he}, online signature verification \cite{lai2021synsig2vec}, person re-id \cite{osnet2022zhou,cdnet2021li,cal2021rao}, and offline writer retrieval \cite{rethis2023icdar}. Apart from Sig2Vec \cite{lai2021synsig2vec} for online signature verification, other models are originally 2D models that take images as input. To adapt these models for our task, we convert them into their 1D version by simply substituting the \emph{Conv2d} module with the \emph{Conv1d} module in Pytorch \cite{pytorch2019paszkes} implementations, which avoids altering their core designs and functionalities. For rigorous evaluation, we repeated each experiment 30 times, randomly selecting the query set and gallery set from the testing data as described in Sec.~\ref{sec::exp_dataset} in each iteration. The results are presented in Table~\ref{Table::sota-comp}, in which we report the average $avg$ and standard deviation $std$ of the performance in the format of $avg$ (\textpm $std$).

\begin{table}[t]
	\renewcommand{\arraystretch}{1.2}
	\caption{Ablation studies on the OLIWER dataset, evaluating the effectiveness of CAIR, HFGA, and Context-Aware FPN (C-FPN).}
	\label{Table::ablation}
	\centering
	\resizebox{0.9\linewidth}{!}{
		\begin{tabular}{c c c c c c c c}
			\whline{1.1pt}
			\textbf{Baseline} & \textbf{CAIR} & \textbf{HFGA} & \textbf{C-FPN} & \textbf{R1} $\uparrow$ & \textbf{R5} $\uparrow$ & \textbf{R10} $\uparrow$ & \textbf{mAP} $\uparrow$\\
			\hline
			\checkmark & ~ & ~ & ~ & 90.78 & 98.56 & 99.14 & 52.71\\
			\hline
			\checkmark & \checkmark & ~ & ~ & 93.95 & 98.56 & 98.85 & 62.44\\
			\hline
			\checkmark & \checkmark & \checkmark & ~ & 94.24 & 97.98 & 99.14 & 64.04\\
			\hline
			\checkmark & ~ & \checkmark & \checkmark & 92.80 & 97.69 & 99.14 & 60.00\\
			\hline
			\checkmark & \checkmark & ~ & \checkmark & 95.68 & 99.42 & 99.71 & 64.90\\
			\hline
			\checkmark & \checkmark & \checkmark & \checkmark & \textbf{96.83} & \textbf{99.71} & \textbf{100.00} & \textbf{68.05}\\
			\whline{1.1pt}
	\end{tabular}}
\end{table}

\begin{table}[t]
	\renewcommand{\arraystretch}{1.2}
	\caption{Ablation study regarding the HFGA block and its variant that utilizes the multi-head attention mechanism to compute the cross-attention, denoted as HFGA-MH.}
	\label{Table::ablation_hfga}
	\centering
	\resizebox{0.75\linewidth}{!}{
		\begin{tabular}{c c c c c}
			\whline{1.1pt}
			\textbf{Variant} & \textbf{R1} $\uparrow$ & \textbf{R5} $\uparrow$ & \textbf{R10} $\uparrow$ & \textbf{mAP} $\uparrow$\\
			\hline
			DOLPHIN w/ HFGA-MH & 95.10 & 99.42 & \textbf{100.00} & 66.55\\
			\hline
			DOLPHIN w/ HFGA & \textbf{96.83} & \textbf{99.71} & \textbf{100.00} & \textbf{68.05}\\
			\whline{1.1pt}
	\end{tabular}}
\end{table}

\begin{table}[!t]
	\renewcommand{\arraystretch}{1.2}
	\caption{Ablation study regarding the vanilla FPN and our introduced Context-Aware FPN (C-FPN).}
	\label{Table::ablation_cfpn}
	\centering
	\resizebox{0.75\linewidth}{!}{
		\begin{tabular}{c c c c c}
			\whline{1.1pt}
			\textbf{Variant} & \textbf{R1} $\uparrow$ & \textbf{R5} $\uparrow$ & \textbf{R10} $\uparrow$ & \textbf{mAP} $\uparrow$\\
			\hline
			DOLPHIN w/ FPN & 95.97 & 98.85 & 99.14 & 67.00\\
			\hline
			DOLPHIN w/ C-FPN & \textbf{96.83} & \textbf{99.71} & \textbf{100.00} & \textbf{68.05}\\
			\whline{1.1pt}
	\end{tabular}}
\end{table}

On four evaluated datasets, it can be observed that DOLPHIN outperforms other methods with substantial margins. From absolute values, mAP proves significantly more challenging than the Ranking accuracies. Despite the greater challenge, our model remarkably surpasses existing approaches in terms of mAP performance. Specifically, on OLIWER and CASIA-OLHWDB2, DOLPHIN surpasses the second-best performers CAL and CDNet by 8.82\% and 7.64\% in mAP, respectively. On DCOH-E and SCUT-COUCH2009, DOLPHIN outperforms the second-best performer Sig2Vec by 7.67\% and 2.20\% in mAP, respectively. DOLPHIN also exhibits improved performance stability, as reflected by lower standard deviations in mAP. In addition, DOLPHIN outshines previous methods across all Ranking metrics. These exceptional results substantiate DOLPHIN’s superior capability in extracting informative writing features, which could be primarily attributed to its effective synergy of temporal and frequency learning.

Furthermore, a noteworthy observation is that domain-agnostic models (marked as \emph{General}) generally exhibit inferior performance compared to other domain-specific models. While they excel in 2D vision tasks such as image classification or object detection, they may not adapt to the 1D sequence modeling manner and often lack tailored designs for the retrieval task. In contrast, other domain-specific methods are dedicatedly designed for either handwriting modeling or information retrieval. For instance, Sig2Vec \cite{lai2021synsig2vec} is specifically crafted to extract effective embeddings for handwritten signatures; CDNet \cite{cdnet2021li}, CAL \cite{cal2021rao}, and OSNet \cite{osnet2022zhou} possess efficient module design for the person retrieval tasks. Therefore, they reasonably yield better performance than the domain-agnostic models. Still, our proposed DOLPHIN outperforms these methods congruently and significantly.

\subsection{Ablation Study}
To evaluate the effectiveness of our proposed modules, we conduct ablation studies on the OLIWER dataset by adding/removing different components. We construct the Baseline by combining the vanilla backbone of MobileNetV2 \cite{mobilev22018sandler} (consisting of the Inverted Residual module), a Selective Pooling layer \cite{lai2021synsig2vec}, and the same \emph{Head} module used in DOLPHIN. Results are summarized in Table~\ref{Table::ablation}.

\textbf{Channel Activation Inverted Residual (CAIR).} To evaluate the impact of the CAIR block, we substitute the Temporal Backbone in DOLPHIN, which comprises the CAIR, with the backbone of MobileNetV2. From line 5 and line 7 in Table~\ref{Table::ablation}, it can be observed that incorporating the CAIR block brings significant improvements in both Rank-1 accuracy and mAP, with a remarkable 8.05\% increase in mAP. The CAIR block is specifically devised to activate the channel information flow and reduce channel redundancy. These results strongly substantiate CAIR's efficacy in empowering channel modeling, consequently enhancing the model performance.

\textbf{High Frequency Gated Attention (HFGA).} The efficacy of the HFGA block is assessed in multiple scenarios. (1) Upon the utilization of CAIR, we integrate the Frequency Network composed of HFGA blocks into the model. This leads to a 1.6\% enhancement in mAP as shown in line 3 and line 4. (2) Upon the utilization of both CAIR and C-FPN, we further add HFGA into the model and form the holistic DOLPHIN model. From line 7, this integration achieves the optimal performance, resulting in a 1.15\% improvement in R1 accuracy and a substantial 3.15\% improvement in mAP. (3) We investigate a variant of the HFGA block that adopts the multi-head attention mechanism \cite{attention2017vaswani} instead of the original single-head attention mechanism, denoted as HFGA-MH. The results are presented in Table~\ref{Table::ablation_hfga}. Interestingly, adopting multi-head attention leads to performance degradation, with the added drawback of consuming $3\times$ GPU memory for training the same architecture model. Therefore, we utilize single-head attention for better performance and computational efficiency. The above experiments underscore the effectiveness of the proposed HFGA block, which could be attributed to HFGA's synergy between temporal and frequency features via the gated cross-attention mechanism. By attending to the informative high-frequency components and facilitating interaction between the temporal and frequency domains, HFGA distills a robust understanding of the writer's unique stylistic traits, thus enhancing the model's performance.

\textbf{Context-Aware FPN.} We evaluate the effectiveness of the Context-Aware FPN by comparing the DOLPHIN model with and without this module. As demonstrated in line 4 and line 7 of Table~\ref{Table::ablation}, this module boosts performance by 2.59\% and 4.01\% in R1 and mAP, respectively. In addition, since we integrate the Global Context (GC) block and FPN to enhance context understanding, we dissect its impact in Table~\ref{Table::ablation_cfpn}. Augmenting the vanilla FPN with the GC block results in performance gains of 0.86\% and 1.05\% in R1 and mAP, underscoring this module's prowess in infusing the model with rich contextual features. By combining FPN and the GC block, the Context-Aware FPN not only captures multi-scale handwriting details but also effectively learns contextual writing styles, thereby bolstering the overall performance.

\subsection{Model Efficiency}
\begin{table}[t]
	\renewcommand{\arraystretch}{1.2}
	\caption{Comparison of the model parameters and FLOPs between DOLPHIN and other methods. We remove the classification layer during parameter calculation. We utilize one online sample with shape $\mathbb{R}^{1000\times14}$ as all models' input for fair comparison on FLOPs, in which 1000 is the length of online handwritten data and 14 is the input feature dimension.}
	\label{Table::comp-param}
	\centering
	\resizebox{0.9\linewidth}{!}{
		\begin{tabular}{c c c c c}
			\whline{1.1pt}
			\textbf{Model} & \textbf{Venue} & \textbf{Domain} & \textbf{Params/M} & \textbf{FLOPs/M}\\
			\hline
			ResNet-34 \cite{resnet2016he} & CVPR'16 & General & 7.22 & 358\\
			MobileNetV2 \cite{mobilev22018sandler} & CVPR'18 & General & 2.18 & 96\\
			EfficientNet-b0 \cite{effnet2019tan} & ICML'19 & General & 7.65 & 159\\
			ConvNeXt-Tiny \cite{convnext2022liu} & CVPR'22 & General & 26.77 & 1300\\
			HorNet-Tiny \cite{hornet2022rao} & NeurIPS'22 & General & 21.07 & 1160\\
			FasterNet \cite{fasternet2023chen} & CVPR'23 & General & 12.01 & 544\\
			GR-RNN \cite{grrnn2021he} & PR'21 & Off-WID & 3.22 & 381\\
			Sig2Vec \cite{lai2021synsig2vec} & TPAMI'22 & On-SV & 0.68 & 144\\
			CDNet \cite{cdnet2021li} & CVPR'21 & Re-ID & 2.21 & 189\\
			CAL \cite{cal2021rao} & ICCV'21 & Re-ID & 53.79 & 2270\\
			OSNet \cite{osnet2022zhou} & TPAMI'22 & Re-ID & 2.13 & 199\\
			NetRVLAD \cite{rethis2023icdar} & ICDAR'23 & Off-WR & 0.30 & 98\\
			\hline
			DOLPHIN (\textbf{Ours}) & This work & On-WR & 2.14 & 361\\
			\whline{1.1pt}
	\end{tabular}}
\end{table}

We conducted a comparative analysis between DOLPHIN and existing models regarding model size and computational efficiency, as illustrated in Table~\ref{Table::comp-param}. We adopt the parameter number and floating point operations (FLOPs) as metrics to assess model size and computational cost, respectively. In the open-set scenario, where merely feature vector output by the model is used for retrieval (Sec.~\ref{sec::metric}), we exclude the final classification layer of all models during metric computation, since it is engaged in only training but not inference. To compute FLOPs, we utilize an online handwriting sample with shape $\mathbb{R}^{1000\times14}$ as all model's input, where 1000 is the length of the online sequence and 14 is the input feature dimension (as detailed in Sec.~\ref{sec::preprocessing}). 

As shown in Table~\ref{Table::comp-param}, the proposed DOLPHIN comprises only 2.14 million parameters, which are even slightly fewer than the classical lightweight model MobileNetV2. The FLOPs of DOLPHIN also stay at a low level, comparable to the small backbone ResNet-34. On a machine equipped with an \emph{Intel(R) Core(TM) i5-8600K @ 3.60GHz} CPU and an RTX 3090 GPU, DOLPHIN processes a sample of shape $\mathbb{R}^{1000\times14}$ in approximately 9.95ms. The modest model size and low inference time underscore DOLPHIN's storage and computation efficiency. Notably, despite having a similar parameter volume and computational load as other lightweight models, DOLPHIN significantly outperforms them, especially in terms of mAP, further emphasizing its effectiveness.

\subsection{Online Writer Identification}
Given a similar functionality of determining the authorship of handwriting between writer retrieval and writer identification, we conduct additional comparisons on the online writer identification task using the OLIWER dataset. Departing from the open-set scenario employed in writer retrieval, we adopt the commonly used closed-set configuration \cite{chen2021level,rstc2022zhang,grrnn2021he} for this task, in which the training and testing samples are from the same writers as opposed to the non-overlapping writer sets in open-set retrieval. We assign 80\% of the total samples of each writer for training and use the rest 20\% for testing, resulting in 538,578/135,439 training/testing samples from the same 1,731 writers. The performance is evaluated using Rank-1, Rank-5, and Rank-10 accuracies \cite{chen2021level,rstc2022zhang,grrnn2021he}, equaling to the CMC curves mentioned in Sec.~\ref{sec::metric}. Experimental results are presented in Table~\ref{Table::comp-wid}. Notably, even in the writer identification task with a distinct setup, DOLPHIN demonstrates superior performance compared to existing methods, further validating its effectiveness and generalizability.

It is worth mentioning that writer retrieval focuses on retrieving unseen writers, while writer identification aims to identify a known set of individuals. Each task caters to different dimensions of applications with context-dependent importance. Our results suggest that DOLPHIN's robust performance extends to both realms, providing a versatile solution for various handwriting analysis challenges.

\begin{table}[t]
	\renewcommand{\arraystretch}{1.2}
	\caption{Comparison between DOLPHIN and existing methods on the online writer identification task using the OLIWER dataset.}
	\label{Table::comp-wid}
	\centering
	\resizebox{0.9\linewidth}{!}{
		\begin{tabular}{c c c c c c}
			\whline{1.1pt}
			\textbf{Model} &  \textbf{Venue} & \textbf{Domain} & \textbf{R1} $\uparrow$ & \textbf{R5} $\uparrow$ & \textbf{R10} $\uparrow$\\
			\hline
			ResNet-34 \cite{resnet2016he} & CVPR'16 & General & 91.28 & 97.83 & 98.73\\
			MobileNetV2 \cite{mobilev22018sandler} & CVPR'18 & General & 86.80 & 96.83 & 98.31\\
			EfficientNet-b0 \cite{effnet2019tan} & ICML'19 & General & 81.54 & 94.27 & 96.56\\
			ConvNeXt-Tiny \cite{convnext2022liu} & CVPR'22 & General & 90.75 & 97.46 & 98.46\\
			HorNet-Tiny \cite{hornet2022rao} & NeurIPS'22 & General & 89.90 & 97.38 & 98.39\\
			FasterNet \cite{fasternet2023chen} & CVPR'23 & General & 65.18 & 80.93 & 85.39\\
			GR-RNN \cite{grrnn2021he} & PR'21 & Off-WID & 91.48 & 97.81 & 98.65\\
			Sig2Vec \cite{lai2021synsig2vec} & TPAMI'22 & On-SV & \underline{94.14} & 98.31 & 98.89\\
			CDNet \cite{cdnet2021li} & CVPR'21 & Re-ID & 92.85 & 97.98 & 98.74\\
			CAL \cite{cal2021rao} & ICCV'21 & Re-ID & 93.93 & 97.80 & 98.35\\
			OSNet \cite{osnet2022zhou} & TPAMI'22 & Re-ID & 93.66 & \underline{98.40} & \underline{99.01}\\
			NetRVLAD \cite{rethis2023icdar} & ICDAR'23 & Off-WR & 79.26 & 94.16 & 96.73\\
			\hline
			DOLPHIN (\textbf{Ours}) & This work & On-WR & \textbf{96.12} & \textbf{99.24} & \textbf{99.58}\\
			\whline{1.1pt}
	\end{tabular}}
\end{table}

\subsection{Cross-Domain Writer Retrieval}
\begin{table}[t]
	\renewcommand{\arraystretch}{1.2}
	\caption{Cross-domain evaluation on the segmented CASIA-OLHWDB2 \cite{casiaolhwdb2011liu} and segmented DCOH-E \cite{DCOH2024dlvc} datasets. A$\rightarrow$B denotes training on the training set of dataset A while testing on the test set of dataset B. We select several well-performing models for comparison.}
	\label{Table::cross}
	\centering
	\resizebox{\linewidth}{!}{
		\begin{tabular}{c c c c c c | c c c c}
			\whline{1.1pt}
			\vspace{-2pt}
			\multirow{2.4}{*}{\textbf{Model}} & \multirow{2.4}{*}{\textbf{Domain}} & \multicolumn{4}{c}{\textbf{OLHWDB2 $\rightarrow$ DCOH-E}} & \multicolumn{4}{c}{\textbf{DCOH-E $\rightarrow$ OLHWDB2}}\\
			\cmidrule(r){3-6}\cmidrule{7-10}
			~ & ~ & \textbf{R1} $\uparrow$ & \textbf{R5} $\uparrow$ & \textbf{R10} $\uparrow$ & \textbf{mAP} $\uparrow$ & \textbf{R1} $\uparrow$ & \textbf{R5} $\uparrow$ & \textbf{R10} $\uparrow$ & \textbf{mAP} $\uparrow$\\
			\hline
			Sig2Vec \cite{lai2021synsig2vec} & On-SV & \textbf{62.28} & \textbf{84.21} & \textbf{89.47} & \textbf{24.55} & 26.96 & 48.53 & 58.82 & 4.95\\
			CDNet \cite{cdnet2021li} & Re-ID & 49.12 & 72.81 & 82.46 & 16.78 & 7.84 & 24.02 & 31.86 & 1.73\\
			CAL \cite{cal2021rao} & Re-ID & 46.49 & 71.05 & 82.46 & 16.32 & 8.33 & 22.55 & 33.33 & 1.74\\
			OSNet \cite{osnet2022zhou} & Re-ID & 39.47 & 69.30 & 77.19 & 17.16 & 11.76 & 24.51 & 33.33 & 1.83\\
			\hline
			DOLPHIN (\textbf{Ours}) & On-WR & 57.02 & 77.19 & 81.58 & 21.26 & \textbf{32.84} & \textbf{57.35} & \textbf{66.67} & \textbf{5.35}\\
			\whline{1.1pt}
	\end{tabular}}
\end{table}
\unskip

\begin{table}[t]
	\renewcommand{\arraystretch}{1.2}
	\caption{The effect of point sampling frequency. We keep the point sampling frequency of DCOH-E constant to 120Hz, cross-validating it with the 30Hz original data and the resampled 120Hz data of CASIA-OLHWDB2. \emph{P. S. F.} here solely refers to the point sampling frequency of CAISA-OLHWDB2.}
	\label{Table::interp_comp}
	\centering
	\resizebox{0.9\linewidth}{!}{
		\begin{tabular}{c c c c c c c}
			\whline{1.1pt}
			\textbf{P. S. F.} & \textbf{Train} & \textbf{Test} & \textbf{R1} $\uparrow$ & \textbf{R5} $\uparrow$ & \textbf{R10} $\uparrow$ & \textbf{mAP} $\uparrow$\\
			\hline
			30Hz & DCOH-E & OLHWDB2 & 4.41 & 12.25 & 18.63 & 0.80\\
			\hline
			120Hz & DCOH-E & OLHDWB2 & 32.84 \textcolor{darkgreen}{$\uparrow$} & 57.35 \textcolor{darkgreen}{$\uparrow$} & 66.67 \textcolor{darkgreen}{$\uparrow$} & 5.35 \textcolor{darkgreen}{$\uparrow$}\\
			\hline
			30Hz & OLHWDB2 & DCOH-E & 39.47 & 65.79 & 71.93 & 11.07\\
			\hline
			120Hz & OLHWDB2 & DCOH-E & 57.02 \textcolor{darkgreen}{$\uparrow$} & 77.19 \textcolor{darkgreen}{$\uparrow$} & 81.58 \textcolor{darkgreen}{$\uparrow$} & 21.26 \textcolor{darkgreen}{$\uparrow$}\\
			\hline
			30Hz & OLHWDB2 & OLHWDB2 & 83.82  & 94.12 & 95.10 & 36.07\\
			\hline
			120Hz & OLHWDB2 & OLHWDB2 & 97.06 \textcolor{darkgreen}{$\uparrow$} & 99.51 \textcolor{darkgreen}{$\uparrow$} & 99.51 \textcolor{darkgreen}{$\uparrow$} & 69.94 \textcolor{darkgreen}{$\uparrow$}\\
			\hline
			\whline{1.1pt}
	\end{tabular}}
\end{table}
\unskip

\begin{table}[t]
	\renewcommand{\arraystretch}{1.1}
	\caption{The effect of pressure information of online handwritten data. We respectively keep the pressure information of the DCOH-E dataset as the original value (denoted as $\mathbbm{X}$) and set it to 1 (denoted as $\mathbbm{1}$), and cross-validate it with the CASIA-OLHWDB2 dataset.}
	\label{Table::pressure_comp}
	\centering
	\resizebox{\linewidth}{!}{
		\begin{tabular}{c c c c c c c}
			\whline{1.1pt}
			\textbf{Pressure} & \textbf{Train} & \textbf{Test} & \textbf{R1} $\uparrow$ & \textbf{R5} $\uparrow$ & \textbf{R10} $\uparrow$ & \textbf{mAP} $\uparrow$\\
			\hline
			$\mathbbm{X}$ & DCOH-E & OLHDWB2 & 32.84 & 57.35 & 66.67 & 5.35\\
			\hline
			$\mathbbm{1}$ & DCOH-E & OLHWDB2 & 36.76 \textcolor{darkgreen}{$\uparrow$} & 63.73 \textcolor{darkgreen}{$\uparrow$} & 73.53 \textcolor{darkgreen}{$\uparrow$} & 6.45 \textcolor{darkgreen}{$\uparrow$}\\
			\hline
			$\mathbbm{X}$ & OLHWDB2 & DCOH-E & 57.02 & 77.19 & 81.58 & 21.26\\
			\hline
			$\mathbbm{1}$ & OLHWDB2 & DCOH-E & 60.53 \textcolor{darkgreen}{$\uparrow$} & 78.07 \textcolor{darkgreen}{$\uparrow$} & 85.96 \textcolor{darkgreen}{$\uparrow$} & 21.04 \textcolor{red}{$\downarrow$}\\
			\hline
			$\mathbbm{X}$ & DCOH-E & DCOH-E & 92.98 & 99.12 & 100.00 & 63.87\\
			\hline
			$\mathbbm{1}$ & DCOH-E & DCOH-E & 86.84 \textcolor{red}{$\downarrow$} & 98.25 \textcolor{red}{$\downarrow$} & 100.00 - & 53.37 \textcolor{red}{$\downarrow$}\\
			\whline{1.1pt}
	\end{tabular}}
\end{table}
\unskip

In real-world retrieval, handwritten data may originate from different regions or acquisition periods. They can be viewed as different data domains with entirely irrelevant distributions. The substantial distributional discrepancies across different domains engender significant cross-domain challenges for the retrieval systems, requiring robust model generalizability. Hence, we attempt to study the model's ability regarding cross-domain generalization by training it on one dataset and directly testing it on another dataset with unseen samples and writers. CASIA-OLHWDB2 \cite{casiaolhwdb2011liu} and DCOH-E \cite{DCOH2024dlvc} have non-overlapping writers and exhibit dramatic distribution differences. Therefore, we use the segmented CASIA-OLHWDB2 and segmented DCOH-E to perform cross-domain evaluations. We experiment with OLHWDB2 $\rightarrow$ DCOH-E and DCOH-E $\rightarrow$ OLHWDB2, in which OLHWDB2 $\rightarrow$ DCOH-E represents training the model on the training set of OLHWDB2 and testing it on the test set of DCOH-E, and vice versa.

The results are listed in Table~\ref{Table::cross}. For OLHWDB2 $\rightarrow$ DCOH-E, Sig2Vec achieves 62.28\% for R1 and 24.5\% for mAP, while DOLPHIN trails behind Sig2Vec achieving 57.02\% and 21.26\% for R1 and mAP. In the DCOH-E $\rightarrow$ OLHWDB2 scenario, DOLPHIN yields the best performance compared to other models, achieving 32.84\% and 5.35\% for R1 and mAP. However, compared to the intra-domain results of Table~\ref{Table::sota-comp}, where the optimal outcomes are 97.06\% and 69.94\% (R1 and mAP) on CASIA-OLHWDB2 and 92.98\% and 63.87\% on DCOH-E, all models' performances on foreign domains consistently suffer from severe degradation. This presents the giant feature gap between handwriting from different domains (datasets), potentially deriving from the difference regarding writing instruments, writing content, and writer conditions. In practice, the disparity in handwriting characteristics could be even more pronounced, posing larger difficulties of cross-domain adaptation for retrieval systems.

We intend to investigate the factors that contribute to this significant gap between different data domains. There are two apparent distinctions in the data properties of CASIA-OLHWDB2 and DCOH-E. (1) Point sampling frequency: DCOH-E exhibits a higher data point sampling frequency during data acquisition of 120Hz, while the frequency of CAISA-OLHWDB2 is 30Hz. (2) Availability of pressure information: DCOH-E provides the pressure information of each handwritten point, while CASIA-OLHWDB2 inherently lacks this feature. Therefore, we speculate that the discrepancies in these two aspects account for the large data gap. We conduct two exploratory experiments to further verify the hypothesis.

First, to explore the influence of point sampling frequency, we sample the data of CASIA-OLHWDB2 at different frequencies and conducted cross-validation with DCOH-E. The comparative results are detailed in Table~\ref{Table::interp_comp}. From lines 2-5, we observe an obvious improvement in cross-domain retrieval performance when increasing the point sampling frequency of CASIA-OLHWDB2 from 30Hz to 120Hz. Particularly, R1 increases from 4.41\% to 32.84\% in lines 2-3 and mAP rises from 11.07\% to 21.26\% in lines 4-5. This suggests that increasing the information density by upsampling data points of handwritten data could help mitigate the distributional gap. Comparing lines 6-7, increasing point sampling frequency could also improve the intra-dataset performance, showing a 33.87\% increment in mAP. This aligns with our operation of resampling all data to 120Hz, given the point sampling frequencies of most datasets are inconsistent and below 120Hz.

Second, we study the effect of writing pressure. To isolate the effect of pressure information in the DCOH-E dataset, we set it to a constant value of 1 to align with CASIA-OLHWDB2 and conduct cross-dataset comparisons. Results are presented in Table~\ref{Table::pressure_comp}. From lines 2-5, it is observed that setting pressure to 1 generally improves cross-domain performances. This is attributed to the alignment of pressure information of both datasets, which minimizes their feature discrepancies. On the other hand, in lines 6-7, we find that the performance on the DCOH-E dataset significantly deteriorates when the pressure information is removed, in which R1 drops 6.14\% and mAP drops 10.5\%. It implies the importance of pressure as a distinguishing factor in online handwriting, especially in the context of online writer retrieval. In addition, it is worth mentioning that all methods benefit from the pressure features as they uniformly take pressure as input (Sec.~\ref{sec::preprocessing}), further underscoring its effectiveness.

These two experiments highlight the significance of aligning properties between two distributionally distinct datasets in bridging their gap. Increasing the inherent alignment of handwritten features, such as ensuring consistent point sampling frequency and setting uniform pressure information, could significantly reduce the disparity between two data domains. Furthermore, the second experiment highlights the importance of pressure information in online handwritten data and its effectiveness in boosting online writer retrieval performance. Given the easy accessibility of pressure data from modern electronic devices, it is advisable to include pressure data collection when constructing new online handwriting datasets.

\section{Conclusion}
\label{sec::conclusion}
In this paper, we propose DOLPHIN, a novel retrieval model aiming to enhance stylistic handwriting representation. DOLPHIN achieves synergistic temporal-frequency interaction with the core idea of High Frequency Gated Attention (HFGA) and Channel Activation Inverted Residual (CAIR). Leveraging the Discrete Wavelet Transform, HFGA extracts high-frequency components of handwriting and conducts gated cross-attention to amplify local writing traits. CAIR is responsible for temporal feature learning, which effectively reduces channel information redundancy via progressive channel interaction. We also introduce the Context-Aware FPN for contextual multi-scale feature fusion. By integrating these three designs, DOLPHIN exhibits exceptional prowess in holistic feature modeling. In addition, we introduce the first large-scale dataset OLIWER for online writer retrieval. OLIWER is constructed by aggregating three online handwriting datasets, containing 674,017 consecutively handwritten phrases from 1,731 writers.

Through comprehensive experiments, we demonstrate the remarkable performance of our proposed DOLPHIN across multiple datasets, which consistently surpasses other domain-agnostic and domain-specific methods by substantial margins. Additionally, we compare DOLPHIN with existing methods on the online writer identification task, further validating its effectiveness and versatility. Furthermore, we explore cross-domain writer retrieval and discover the pivotal role of increasing feature alignment to bridge the feature gap between different handwritten data domains. Higher point sampling frequency of handwriting data and the inclusion of pressure data prove instrumental in enriching writing feature quality, consequently bolstering retrieval precisions.

While this work focuses on using Chinese handwriting for online writer retrieval, DOLPHIN is designed to be language-agnostic. The choice of Chinese datasets is driven by their large scale and availability of character/phrase-level annotations, which collectively enabled the creation of a large-scale, phrase-level online writer retrieval dataset. Public datasets in other languages, such as IAM-OnDB (English) \cite{iamonline2005liwicki}, ADAB (Arabic) \cite{adab2016ijcnn}, HANDS-VNOnDB (Vietnamese) \cite{vietnamese2018icfhr}, and MRG-OHMW (Mongolian) \cite{mongolian2016icpr}, either lack the requisite scale or sufficiently detailed annotations. However, DOLPHIN's design principles and underlying mechanisms show potential for application to other languages, meriting future investigation.

\section*{Acknowledgement}
This research is supported in part by National Key Research and Development Program of China (2022YFC3301703) and National Natural Science Foundation of China (Grant No.: 62441604, 62476093).

\bibliographystyle{IEEEtran}
\bibliography{reference} 

\end{document}